# Finding any Waldo: zero-shot invariant and efficient visual search

Mengmi Zhang[1,2,3], Jiashi Feng[2], Keng Teck Ma[3], Joo Hwee Lim[3], Qi Zhao[4], and Gabriel Kreiman[1*]

[1] Children's Hospital, Harvard Medical School

[2] National University of Singapore

[3] Agency for Science, Technology and Research (A*STAR)

[4] University of Minnesota Twin Cities

[*] To whom correspondence should be addressed: gabriel.kreiman@tch.harvard.edu

Text statistics

Number of figures: 6

Number of supplementary figures: 14

Number of words in abstract: 139

**Abstract**

Searching for a target object in a cluttered scene constitutes a fundamental challenge in daily vision. Visual search must be selective enough to discriminate the target from distractors, invariant to changes in the appearance of the target, efficient to avoid exhaustive exploration of the image, and must generalize to locate novel target objects with zero-shot training. Previous work has focused on searching for perfect matches of a target after extensive category-specific training. Here we show for the first time that humans can efficiently and invariantly search for natural objects in complex scenes. To gain insight into the mechanisms that guide visual search, we propose a biologically inspired computational model that can locate targets without exhaustive sampling and generalize to novel objects. The model provides an approximation to the mechanisms integrating bottom-up and top-down signals during search in natural scenes.

## Introduction

Visual search constitutes a ubiquitous challenge in natural vision, including daily tasks such as looking for the car keys at home. Localizing a target object in a complex scene is also important for many applications including navigation and clinical image analysis. Visual search must fulfill four key properties: (1) selectivity (to distinguish the target from distractors in a cluttered scene), (2) invariance (to localize the target despite changes in its appearance or even in cases when the target appearance is only partially defined), (3) efficiency (to localize the target as fast as possible, without exhaustive sampling), and (4) zero-shot training (to generalize to finding novel targets despite minimal or zero prior exposure to them).

Visual search is a computationally difficult task due to the myriad possible variations of the target and the complexity of the visual scene. Under most visual search conditions, the observer does *not* seek an identical match to the target object at the pixel level. The target object can vary in rotation, scale, color, illumination, occlusion, and other transformations. Additionally, the observer may be looking for any exemplar from a generic category (e.g. looking for any chair, rather than a specific one). Robustness to object transformations has been a fundamental challenge in the development of visual recognition models where it is necessary to identify objects in a way that is largely invariant to pixel-level changes (e.g.[1-8], among many others). The critical constraint of invariance in recognition has led to hierarchical models that progressively build transformation-tolerant features that are useful for selective object identification.

In contrast with the development of such bottom-up recognition models, less attention has been devoted to the problem of invariance in visual search. A large body of behavioral[9-12] and neurophysiological[13-16] visual search experiments has focused on situations that involve identical target search. In those experiments, the exact appearance of the target object is perfectly well defined in each trial (e.g., searching for a tilted red bar, or searching for an identical match to a photograph of car keys). Some investigators have examined the ability to search for faces that are rotated with respect to a canonical viewpoint[17], but there was no ambiguity in the target appearance, therefore circumventing the critical challenge in invariant visual search. Hybrid visual search studies examine situations where the observer is looking for two or more objects, but the appearance of those objects is fixed[18]. Several studies have evaluated reaction times during visual search for generic categories as a function of the number of distractors (e.g.,[19,20]).

Traditional template-matching computational algorithms do not perform well in invariant object recognition. In visual search tasks, template-matching shows selectivity to distinguish an identical target from distractors, but fails to robustly find transformed versions of the target. To circumvent this problem, investigators have developed object detection, object localization, and image retrieval approaches which can successfully and robustly localize objects, at the expense of having to extensively train those models with the sought targets and exhaustively scan the image through sliding windows[21]. To localize objects, recent work focuses on deep neural networks requiring a large amount of supervised data, such as bounding boxes or object segmentations[21,2223]. Typically, these approaches either use a sliding window or propose regions of interest uniformly over a grid, performing feed-forward classification for each region and making decisions about the presence or absence of the target. An analogous strategy is used in image retrieval tasks where a similarity score is computed between a query and each candidate image[24,25]. These heuristic methods are computationally inefficient (in terms of the number of ``fixations'' or proposed regions required to find the target), and require extensive class-specific training.

Most of these computer vision approaches bear no resemblance to the neurophysiological architecture of visual search mechanisms in cortex. In contrast with heuristic algorithms based on sequential scanning and class-specific supervised training, when presented with a visual search task, observers rapidly move their eyes in a task-dependent manner to search for the target, even when the exact appearance of the target is unknown and even after merely single-trial exposure to the target. When presented with an image, and before taking into account any task constraints, certain parts of the image automatically attract attention due to bottom-up saliency effects[26]. Task goals, such as the sought target in a visual search paradigm, influence attention allocation and eye movements at the behavioral[9,12,27,28] and neurophysiological levels[14,15,29,30]. Task-dependent modulation of neurophysiological responses is likely to originate in frontal cortical structures[15,31] projecting in a top-down fashion onto visual cortex structures[29,32]. Several computational models have been developed to describe visual search behavior or the modulation of responses in visual cortex during feature-based attention or visual search (e.g.,[10-12,27,33-39]).

In the current work, we first set out to quantitatively assess human visual search behavior by evaluating selectivity to targets versus distractors, tolerance to target shape changes, and efficiency. We conducted three increasingly more complex tasks where we measured eye movements while subjects searched for target objects in complex scenes. To gain insight into the mechanisms that guide visual search behavior, we developed a biologically inspired computational model, and evaluated the model in terms of the four key properties of visual search using the same images that human observers were exposed to. We show that humans can efficiently locate target objects despite large changes in their appearance and despite having had no prior experience with those objects. The proposed computational model can efficiently localize target objects amidst distractors in complex scenes, can tolerate large changes in the target object appearance, and can generalize to novel objects with no prior exposure. Furthermore, the model provides a first-order approximation to predict human behavior during visual search.

**Results**

We considered the problem of localizing a target object that could appear at any location in a cluttered scene under a variety of shapes, scales, rotations and other transformations. We conducted 3 increasingly more difficult visual search experiments where 45 subjects had to move their eyes to find the target (**Figure 1**). We propose a biologically inspired computational model to account for the fixations during visual search (**Figure 2**). The structure of the three tasks was similar and is schematically illustrated in **Figure 1** (**Methods**).

**Searching for a target within an array of objects**

Many visual search studies have focused on images with multiple isolated objects presented on a uniform background such as the ones in Experiment 1 (**Figure 1A, 3A**). We used segmented grayscale objects from 6 categories from the MSCOCO data set[40] (**Methods**). After fixation, 15 subjects were presented with an image containing a word describing the object category and a target object at a random 2D rotation (**Figure 1A**). The size of the object was 5 degrees of visual angle. After an additional fixation delay, a search image was introduced, containing a different rendering of the target object, randomly located in one of 6 possible positions within a circle, along with 5 distractors from the other categories. The target was always present and appeared only once. The rendering of the target in the search image was different from the one in the target image (e.g., **Figure 3A**): it was a different exemplar from the same category, and it was shown at a different random 2D rotation. Subjects were instructed to rapidly move their eyes to find the target. Example fixation sequences from 5 subjects are shown in **Figure 3C**: in these examples, subjects found the target in 1 to 4 fixations, despite the fact that the rendering of the target in the search image involved a different sheep, shown at a different 2D rotation. The target locations were uniformly distributed over the six possible positions (**Fig. S1A**) and subjects did not show any appreciable location biases since the distribution of fixations was uniform across all 6 locations (**Fig. S1B**). Subjects made their first fixation on average at 287±152 ms (mean±SD, **Figure 3D**). The distribution of fixation latencies for the first 6 saccades is shown in **Fig. S2A**, showing an interval between fixations of 338±203 ms. The rapid deployment of eye movements during visual search is consistent with previous studies[10], and shows that subjects followed the instructions, probably without adopting alternative strategies such as holding fixation in the center and searching for the target purely via covert attention (**Discussion**).

Subjects located the target in 2.60±0.22 fixations (mean±SD, **Figure 3E**), corresponding to 640±498 ms (mean±SD, **Fig. S2B**). The number of fixations required to find the target was significantly below the number expected from a null model implementing random exploration of the 6 possible object locations, which would require 3.5 fixations in this experiment (compare red line versus black dashed line in **Figure 3E**, $p<10^{-15}$, two-tailed t-test, t=10, df=4473)). Even in the first fixation, subjects were already better than expected by chance (performance = 26.4±4.1%). At 6 fixations, the cumulative performance was below 100% (93.3±1.6%), since subjects tended to revisit the same locations in a small fraction of the fixations, even when they were wrong. The appearance of the target in the search image was different from that in the target image: it was a different exemplar from the same category, shown at a different random 2D rotation. The number of fixations required to find the target was lower when the target was identical in both images (**Fig. S3A-B**), yet subjects were able to efficiently and robustly locate the target despite changes in 2D rotation (**Fig. S3B**) and despite the exemplar differences (**Fig. S3A**).

In order to better understand the guidance mechanisms that incorporate target shape information to dictate the sequence of fixations to find the target, we implemented a computational model inspired by neurophysiological recordings in macaque monkeys during visual search tasks. The Invariance Visual Search Network (IVSN) model consists of a deep feed-forward network that mimics processing of features along ventral visual cortex, a way of temporarily storing information about the target tentatively associated with pre-frontal cortex, modulation of visual features in a top-down fashion to generate an attention map, and sequential selection of fixation locations (**Figure 2B,** see **Methods** for details). Of note, the IVSN model was neither trained with any of the images used in the current study, nor was it trained in any way to match human performance. The same images used for the psychophysics experiments were submitted to the model. **Figure 3B** (left) shows the resulting attention map generated by the model in response to the target and search images from **Figure 3A**. From this attention map, the model generated a sequence of fixations and was able to locate the target in 3 fixations (**Figure 3B**, right). Despite the lack of training with this image set, and the large degree of heterogeneity between the shape presented in the target image and the features of the target in the search image, the model was able to efficiently locate the target in 2.80±1.71 fixations across all the trials (**Figure 3E**, blue). The IVSN model performed well above the null chance model ($p<10^{-11}$, two-tailed t-test, t=7.1, df=598), even in the first fixation (performance = 31.6±0.5% compared to chance = 16.7%). The model had infinite inhibition-of-return and therefore never revisited the same location, by construction thus achieving 100% performance at 6 fixations (see further elaboration of this point under "Extensions and variations of the IVSN model", **Fig. S11**, and **Discussion**). Although there were no free parameters in the model tuned to match human visual search, the IVSN model performance was similar to human behavior. The strong resemblance between the IVSN model and human performance shown in **Figure 3E** should *not* be over interpreted: there was still a small difference between the two (p=0.03, two-tailed t-test t=2.2, df=4473); in addition, we will discuss below other differences between humans and the IVSN model. Similar to human behavior, the model required fewer fixations when the rotation of the target object as rendered in the target image matched the one in the search image, but the model was also able to efficiently locate the target at all the rotations tested (**Fig. S3A-B**).

We considered a series of alternative null models to further understand the nature of image features that guide visual search (**Fig. S4A**). In the sliding window model, commonly used in computer vision approaches, a window of a fixed size sequentially scans the image (here scanning was restricted to the 6 objects on the screen), which is equivalent to random search with infinite inhibition of return in this case, and fails to explain human search behavior. Visual search was not driven by pure bottom-up saliency features as represented by the Itti and Koch model[26]. The weight features in the ventral visual cortex part of the model are important to generate the shape-invariant target-dependent visual attention map, as demonstrated by two observations: (i) randomizing those weights led to chance performance (RanWeight model); (ii) template matching algorithms based on pixels, using rotated templates or not, which are poor at invariant visual object recognition, were insufficient to explain human search behavior (Template Matching model). In sum, both humans and the IVSN model significantly outperformed all of these alternative null models.

**Searching for a target in natural scenes**

The object array images used in Experiment 1 lack several critical components of real world visual search. In natural scenes, there is no fixed type and number of distractors equidistantly arranged in a circle, the target object is not segmented nor is it generally present on a uniform background, and the appearance of the target object can vary along multiple dimensions that are not pre-specified. In Experiment 2, we directly tackled visual search in natural images (**Figure 4**). The structure of the task was essentially the same as that in Experiment 1 (**Figure 1B**) with the following differences: (i) search images involved natural images (e.g., **Figure 4A**), (ii) objects and distractors were not restricted to 6 object categories, (iii) the appearance of the target object in the target image could vary along multiple dimensions from that in the search image, (iv) a trial was ended if the target was not found within 20 seconds, and (v) to ensure that the target was correctly found, subjects had to use the computer mouse to click on the target location (**Methods**). The target locations were randomly and approximately uniformly distributed over all the test images (**Fig. S1D**). Subjects made fixations throughout the entire search image, with certain biases such as a larger density of fixations in the center and a smaller density of fixations along the borders (**Fig. S1E**). Similar to Experiment 1, subjects made rapid fixation sequences. **Figure 4C** shows example sequences where subjects were able to rapidly find the target in 2 to 5 fixations despite the changes in target appearance and despite the large amount of clutter in the image. The first fixation occurred at 285±135 ms (**Figure 4D**), and the interval between fixations was 290±197 ms (the distribution for the first 6 fixations is shown in **Fig. S2C**). On average, the last set of fixations became progressively closer to the target (**Fig. S2H**).

Subjects found the target in 1867±2551 ms (**Fig. S2D**), which was about three times as long as in Experiment 1 (**Fig. S2B**).

Subjects located the target in 6.2±0.7 fixations (**Figure 4E,** red). Performance saturated at about 15 fixations, well below 100%. In 16.4±5.9% of the images, subjects were unable to find the target in the 20 seconds allocated per trial, hence human performance was well below ceiling. Human performance was more efficient than the chance model ($p<10^{-15}$, two-tailed t-test, t=14, df=3247). Subjects tended to revisit the same locations even though the target was not there. In part because of this behavior, the null chance model showed a higher cumulative performance after 20 fixations. The average number of fixations that humans required to find the target was below that expected from the null chance model. Even in the first fixation, subjects were better than expected by chance (performance = 18.3±3.8% versus 7.0±0.2%). The target as rendered in the search image could be larger or smaller than the one in the target image. Intuitively, it could be expected that performance might monotonically increase with the target size in the search image. However, subjects performed slightly better when the size of the target in the search image was similar to the original size in the target image. Subjects were still able to robustly find the target across large changes in size (**Fig. S3D**). In addition to size changes, the target's appearance in the search was generally different in many other ways, which we quantified by computing the normalized Euclidian distance between the target in the target image and in the search image. Subjects robustly found the target despite large changes in its appearance (**Fig. S3C**).

Next, we investigated the performance of the IVSN model in natural images. Importantly, we used *exactly the same model* described for Experiment 1, with no additional tuning or any free parameters adjusted for Experiment 2. An example attention map and scanpath for the IVSN model is shown in **Figure 4B** in response to the target and search images from **Figure 4A**: the model located the target object in 3 fixations even though it had never encountered this target or any similar target before, despite the large amount of clutter, and despite the visual appearance changes in the target. The IVSN model was successful in efficiently locating the target object in natural scenes, requiring 8.3±7.5 fixations on average (**Figure 4E**, blue). The IVSN model performed well above the null chance model ($p<10^{-15}$, two-tailed t-test, t=8.5, df=478), even in the first fixation (14±5% versus 7.0±0.2%). In contrast to the discussion in the previous paragraph for humans, the IVSN model had infinite inhibition-of-return never revisiting the same location, and achieving 100% accuracy in about 45 fixations. Humans outperformed the model up to approximately fixation number 10, but the model performed well above humans thereafter. Consistent with human behavior, visual search performance by the IVSN model was also robust to large differences between the size of the target as rendered in the search images and target images (**Fig. S3D**) and it was also robust to other changes in target object appearance (**Fig. S3C**).

As described in Experiment 1, we considered several alternative null models, all of which were found to show lower performance than humans and the IVSN model (**Fig. S4B**). A pure bottom-up saliency model was worse than chance levels, because it did not incorporate features relevant to the target and instead concentrated on regions of high contrast in the image that were not relevant to the task. Similarly, template matching models were also worse than chance because they generated attention maps that emphasized regions that showed high pixel-level similarity to the target without incorporating invariance and therefore failing to account for the transformations in the target object shape present in the search image.

The ventral visual cortex part of the model (VGG16 architecture) was pre-trained on 1000 categories from the ImageNet dataset (**Methods**). Although all of the target objects and images that we used in both Experiment 1 and Experiment 2 were different from those in the ImageNet dataset, the categories of target objects in 100 of the 240 images in Experiment 2 were among the 1000 ImageNet categories. To evaluate whether the IVSN model can generalize to search for target object categories that it has never encountered before, we separately analyzed the 140 target objects from Experiment 2 belonging to categories that are *not* part of

ImageNet (**Methods**). There was a small improvement in performance for the 100 images with ImageNet category targets versus the 140 images with novel category targets but this difference was not statistically significant (**Fig. S5**, p=0.25, two-tailed t-test, t=1.2, df=238). The IVSN model was still able to successfully and efficiently find the target even for object categories that were completely novel, with zero prior experience.

**Searching for Waldo**

The IVSN model could find objects and object categories that it had never encountered before. To further investigate invariant visual search for novel objects, we designed Experiment 3 to test IVSN with more extreme images that bear no resemblance to those used in Experiments 1 and 2, or to the images in the ImageNet data set. In Experiment 3, we considered the traditional "Where is Waldo" task[41] (**Figure 5**). The Waldo images comprise colorful drawings full of clutter with scene statistics that are very different from those in natural images. The structure of Experiment 3 was similar to that of Experiment 2, except that a picture of Waldo was only presented at the beginning of the experiment and the target was not shown in every trial (**Figure 1C**). The target locations were randomly and approximately uniformly distributed over all the test images (**Fig. S1G**). Subjects made fixations throughout the entire search image, with certain biases such as a higher density in the center and a smaller density of fixations along the borders (**Fig. S1H**). Similar to Experiments 1 and 2, subjects made rapid sequences of fixations (examples are shown in **Figure 5C**), with the first fixation occurring at 264±112 ms (**Figure 5D**), and an interval between fixations of 278±214 ms (the distribution for the first 6 fixations is shown in **Fig. S2E**). As described in Experiment 2, on average, subjects progressively became closer to the target in their last set of fixations (**Fig. S2I**).

Searching for Waldo is generally considered to be a difficult challenge for humans, as confirmed by our results. On average, subjects found the target in 6051±4962 ms

(**Fig. S2F**), which was more than three times as long as in Experiment 2 and more than nine times as long as in Experiment 1. It took subjects 21.1±3.1 fixations to find the target (**Figure 5E**). Performance reached a plateau at about 60 fixations, well below 100%. In 26.9±9.6% of the images, subjects were unable to find the target in the 20 seconds allocated per trial. Despite the difficulty of the task and despite the fact that the null chance model had infinite inhibition of return, subjects were able to find Waldo more efficiently than by random exploration ($p<10^{-15}$, two-tailed t-test, t=18, df=800). Similar to the previous two experiments, in Experiment 3, there were also differences between the rendering of the target object in the search image and target image. Subjects were able to find Waldo despite these changes in target object appearance (**Fig. S3E**).

We evaluated the performance of the IVSN model on the images from Experiment 3, without fine-tuning any parameters. The IVSN model had no prior experience with Waldo images or drawings of any kind. An example attention map and scanpath for the IVSN model for the example image in **Figure 5A** is shown in **Figure 5B**: the model located Waldo in 9 fixations. The IVSN model was successful in efficiently locating Waldo, requiring 29.0±21.6 fixations on average (**Figure 5E**, blue). The IVSN model performed well above the null chance model ($p<10^{-15}$, two-tailed t-test, t=10, df=116). Despite the difficulty of the task, humans were more efficient in finding Waldo than the IVSN model (p=0.001, two-tailed t-test, t=3.3, df=784). The performance of the IVSN model was robust to changes in the appearance of the target object (**Fig. S3E**). As described in Experiments 1 and 2, the alternative null models that we considered did not perform as well as humans or the IVSN model (**Fig. S4C**).

**Human search for novel objects**
All the objects presented in Experiments 1 through 3 were novel for the IVSN model. Although the human subjects had never seen the exact same objects in Experiments 1 and 2 before, they had extensive prior experience with similar objects from the same categories. Additionally, all human subjects had experience with the Waldo character in Experiment 3. In order to assess whether human subjects are able to search for novel objects that they have never encountered before, we conducted an additional experiment using novel objects such as the ones shown in **Fig. S10A** (**Methods**). The structure of the task (**Fig. S10B**) was similar to the one in Experiment 1 (**Fig. 1A**), except that the category name was not included. In addition to the trials with novel objects, other randomly interleaved trials included the same objects from Experiment 1 (known objects) for direct comparison with the same subjects. To ensure a fair comparison, we matched the difficulty of the task for novel objects and known objects by making the distribution of target - distractor similarity for novel objects close to the corresponding distribution for known objects (**Fig. S10C**). Humans were able to efficiently find novel objects, with a performance above that expected by chance (**Fig. S10D**, novel objects: 2.42±1.43 fixations, $p<10^{-15}$, t=13, df=2361; known objects: 2.54±1.42 fixations, $p<10^{-15}$, t=12, df=3515). Average performance for novel objects was slightly above performance for known objects (p=0.004, t=2.9, df=5278, two-tailed t-test), but this difference was small and might potentially be attributable to small differences in task difficulty despite our attempts to match the two. We conclude that human subjects are capable of searching for novel objects that they have never encountered before. As expected based on the results in Experiment 1, the IVSN model was also able to efficiently locate the known and novel objects in this experiment (**Fig. S10E**).

**Image-by-image comparisons**

The results presented thus far compared *average* performance between humans and models considering *all* images. We next examined consistency in the responses at the image-by-image level. For a given image, the IVSN model goes through a sequence of fixations to find the target (e.g., **Figures 3B, 4B, 5B**) and each subject traverses his/her own sequence of fixations towards the target (e.g., **Figures 3C, 4C** and **5C**). We considered different metrics to compare those fixation sequences (**Fig. S6, Methods**).

We started by considering the total number of fixations required to find the target in each image in Experiment 1. First, we evaluated whether subjects would produce a consistent number fixations for the exact same visual search problem. Unbeknown to the subjects, some of the exact same target images and search images were repeated (intermixed in random order). These repetitions allowed us to evaluate the degree of within-subject consistency. The correlation coefficient in the number of fixations required to find the target between the first instance and repeated instance of the same images ranged from 0.17 to 0.45 for the 15 subjects (0.31±0.09, **Fig. S7D**). In other words, there was a significant degree of variability in each individual subject's number of fixations in response to the exact same trial. This definition of within-subject consistency assumes that subjects have no memory of the first instance of the same image. In an extreme case, if subjects had perfect memory of the target location in every image, we would expect that they could rapidly find the target in the second instance of the search image. However, none of the subjects revealed such strong memory effects, which would be evident as increased values below the diagonal in the boxed matrices in **Fig. S7D1**. When considering overall performance, there was almost no difference between the first and second instances of each image (two-tailed t-tests: Exp1, p=0.96 t=0.06 df=8357; Exp2, p=0.28 t=1.1 df=6011; Exp3, p=0.29 t=1.1 df=1454). Next, we compared whether different subjects required the same number of fixations to find the target. The correlation in the number of fixations between subjects ranged from -0.03 to 0.38 (0.21±0.09, **Fig. S7D2**). When comparing the IVSN model to humans, the correlations ranged from -0.05 to 0.12 (0.03±0.05, **Fig. S7D3**). Thus, even when the overall performance of the IVSN model was very similar to that of human subjects (**Figure 3E**), there were many images that were easy for humans and hard for the model, and vice versa (see **Fig. S7A** for several individual image examples). Although there was generally a large degree of variability, subjects were slightly more consistent with themselves than with other subjects, and the between-subject consistency was slightly higher than the consistency with the IVSN model. These conclusions also extend to Experiments 2 and 3 (**Fig. S7**). In both experiments, there was a small, but significant, correlation between the IVSN model and human subjects, which was also slightly smaller than the between-subject correlation.

The number of fixations provides a summary of the efficacy of visual search but does not capture the detailed spatiotemporal sequence guiding search behavior, as illustrated in **Fig. S6**. For example, both sequence number 2 and sequence number 5 in **Fig. S6** differ from the primary scanpath by one fixation, yet the former captures the search behavior in the primary scanpath much better. We used the scanpath similarity score, defined by Borji and colleagues[27], to compare two fixation sequences. This is a metric derived from comparisons of DNA sequences and captures the spatial distance between saccades in two sequences as well as their temporal evolution. The similarity score ranges from 0 (maximally different sequences) to 1 (identical sequences). **Figure S6** provides examples of the scanpath similarity score for pairwise comparisons of several sequences. As described in the previous paragraph, we first compared the scanpath similarity score for repetitions of identical targets and search images within a subject, followed by a comparison between subjects and finally a comparison between the IVSN model and the subject, the results are summarized in **Figure 6**. For a given fixation sequence length *x*, we compared the first x fixations for all images that had at least *x* fixations. As noted in the previous paragraph when comparing the number of fixations, within-subject comparisons yielded slightly more similar sequences than between-subject sequences in all 3 experiments ($p<10^{-9}$). The between-subject scanpath similarity scores, in turn, were higher than the IVSN-human similarity scores for all 3 experiments. The IVSN-human similarity scores were higher than the human-chance similarity scores for all 3 experiments. Similar conclusions were reached when comparing all sequences irrespective of their length (Fig. S8), except that the average scanpath similarity score for IVSN-model comparisons was not statistically significant in Experiment 3.

In sum, the best predictor for a given subject's pattern of fixations during visual search is his/her own previous behavior under identical circumstances, followed by another human subject's behavior in an identical trial, followed by the IVSN model. The IVSN model was able to capture human eye movement behavior at the individual image level in terms of the number of fixations as well as the spatiotemporal pattern of fixations.

**Extensions and variations to the IVSN computational model**

In this section, we consider variations of the IVSN model architecture and revisit several simplifications and assumptions of the model.

All the results presented thus far assume that the model can perfectly recognize whether the target is present or not at the fixated location. After each fixation, an "oracle" decides whether the target is present and therefore the trial should be terminated or to continue searching and move on to a new location. Rapidly recognizing whether the target is present or not is not easy, particularly in Experiments 2 and 3. Human subjects sometimes fixated on the target, yet failed to recognize it, and continued the search process (**Fig. S12A-B**). Examples of this behavior are illustrated for Subjects 1 and 5 in **Figure 4C** – although it is difficult to appreciate this effect because of the small figure size – where the second fixations land on the target, yet the subjects make additional saccades and subsequently return to the target location. For fair comparison, all the human psychophysics results presented thus far also used an oracle for recognition (search was considered to be successful the first time that a fixation landed on the target). Without the oracle, human performance was lower but still well above chance (Experiment 2: $p<10^{-15}$, $t=14$, $df=3247$, **Fig. S12C**; Experiment 3: $p<10^{-15}$, $t=18$, $df=800$, **Fig. S12D**). We introduced a simple recognition component into the model to detect whether the target was present or not based on the features of the object at the fixated location (IVSN$_{recognition}$, **Fig. S11A-C**, **Methods**). As expected, the IVSN$_{recognition}$ model's performance was slightly but not significantly below that of IVSN, particularly in the most challenging case of Experiment 2, but it was still able to find the target well above chance levels (Experiment 1: $p<10^{-11}$, $t=7.3$, $df=594$,

**Fig. S11A**; Experiment 2: $p<10^{-13}$, t=8, df=434, **Fig. S11B**; Experiment 3: $p<10^{-15}$, t=12, df=112, **Fig. S11C**).

Another assumption in the IVSN model is that it has infinite memory of fixated locations and never returns to a previously visited location (infinite inhibition of return). In contrast, humans show a finite memory and tend to revisit the same locations not only for the target (**Fig. S12C-D**) but also for other non-target locations (e.g. subject 1 in **Figure 5C**), as reported in previous work [42]. We fitted an empirical function to describe the probability that subjects would revisit a location at fixation $i$ given that they had visited the same location at fixation $j<i$[42]. We incorporated this empirical function into the IVSN model so that previous fixated locations could be probabilistically revisited, thus creating a model with finite inhibition of return (IVSN$_{fIOR}$, **Methods**, **Fig. S11D-F**). The IVSN$_{fIOR}$ model showed lower performance than the IVSN model but this difference was not significant or marginally significant (Experiment 1: p=0.11; Experiment 2: p=0.02; Experiment 3: p=0.07): this is expected in the current implementation based on an oracle recognition system because the search ended whenever the model fixated on the target. Despite this drop in performance, the IVSN$_{fIOR}$ model was still able to find the target better than chance (Experiment 1: $p=10^{-15}$, t=9.7, df=864; Experiment 2: $p<10^{-15}$, t=11, df=617; Experiment 3: $p=10^{-15}$, t=16, df=145, two-tailed t-tests). Furthermore, the performance of the IVSN$_{fIOR}$ model was closer to human performance for all 3 experiments (**Fig. S11D-F**, IVSN$_{fIOR}$ versus human performance: Experiment 1: p=0.87; Experiment 2: p=0.03; Experiment 3: p=0.29; two-tailed t-tests).

Another difference between humans and the model is the size of saccades (**Fig. S11G-I**). For example, in Experiment 2, the average saccade size was 7.6±5.7 degrees for humans and 16.8±8.4 degrees for IVSN (**Fig. S11H**, $p<10^{-15}$, two-tailed t-test, t=62, df=22960). The human saccade sizes were also different from those in the IVSN model in Experiment 3 (**Fig. S11I**, $p<10^{-15}$, two-tailed t-test, t=100 df=29263). Humans typically made relatively small saccades, following a distribution that resembles a gamma distribution (**Fig. S11H-I**). In contrast, the saccade sizes for the model were more or less uniformly distributed (**Fig. S11H-I**). We used the empirical distribution of saccade sizes to probabilistically constrain the saccade sizes for the model, creating a new variation of the model, IVSN$_{size}$ (**Methods**) The distribution of saccade sizes for the IVSN$_{size}$ model resembled those of humans. IVSN$_{size}$ showed similar performance to IVSN (Experiment 1: p=0.97; Experiment 2: p=0.52; Experiment 3: p=0.47; **Fig. S11J-L**), suggesting that the distribution of saccade sizes plays a lesser role in overall search efficiency.

Attentional modulation based on the target features is implemented in the IVSN model as a top-down signal from layer 31 to layer 30 in the VGG16 architecture (Fig. 2, **Methods**). Connectivity in cortex is characterized by ubiquitous top-down signals at *every* level of the ventral visual stream. We considered variations of the model where attention modulation was implemented via top-down signaling at different levels: layer 31 to 30 (default, as described in **Fig. 2**), layer 24 to 23 (IVSN$_{24 \rightarrow 23}$), layer 17 to 16 (IVSN$_{17 \rightarrow 16}$), layer 10 to 9 (IVSN$_{10 \rightarrow 9}$), layer 5 to 4 (IVSN$_{5 \rightarrow 4}$) (**Fig. S13**). In general, these model variations were also able to find the target above chance levels (all models were statistically different from chance except for IVSN$_{5 \rightarrow 4}$ in Experiment 1). The low-level features (layer 5 to layer 4) showed the lowest performance, probably because they lack the degree of invariance that is built along the ventral stream hierarchy and required to search for objects whose shape is different in the target and search images. Generally, model features at higher levels showed better performance but the trend was not monotonic. For example, IVSN$_{24 \rightarrow 23}$ showed slightly better performance than IVSN in Experiment 1 (**Fig. S13A**), but this difference was not statistically significant (p=0.045, two-tailed t-test, t=2, df=299).

We used the VGG16 architecture as an approximation to ventral visual cortex to extract visual features from the target image and the search image in the IVSN model (**Figure 2**). There are multiple alternative, yet conceptually similar, deep convolutional architectures that have been used for visual object recognition including AlexNet[5], ResNet[43] and FastRCNN[21]. In **Fig. S14**, we report results obtained by replacing the VGG16 visual cortex part of the model by one of those other alternative architectures creating $IVSN_{AlexNet}$, $IVSN_{ResNet}$, and $IVSN_{FastRCNN}$ (**Methods**). The performance of all of these models was above chance in all the experiments (p<0.006). Overall, the performance of these alternative architectures was similar to that of IVSN but some of them yielded a statistically significant difference with IVSN: $IVSN_{Alexnet}$: p<0.01 in Experiment 1; $IVSN_{ResNet}$: $p<10^{-7}$ in Experiment 2 (**Fig. S14**).

**Discussion**

We examined 219,601 fixations to evaluate how humans search for a target object in a complex image under approximately realistic conditions and proposed a biologically plausible computational model that captures essential aspects of human visual search behavior. We investigated three progressively more complex visual search scenarios and found that subjects were able to efficiently locate the target in object arrays (**Figure 3**), in natural images (**Figure 4**) and in Waldo images (**Figure 5**) despite large changes in the appearance of the target object when rendered within the search image. Visual search behavior could be approximated by a neurophysiology-inspired computational network consisting of a bottom-up architecture resembling ventral visual cortex, a pre-frontal cortex-like mechanism to store the target information in working memory and provide top-down guidance for visual search, and a winner-take-all and inhibition-of-return mechanism to direct fixations. Both humans and the Invariant Visual Search Network (IVSN) model, demonstrated selectivity, efficiency and invariance, and did not require any training whatsoever with the sought targets.

Human visual search was efficient in that it required fewer fixations than alternative null models including random search, template matching, and sliding window models (**Figures 3E, 4E, 5E**). In other words, humans deliberately and actively sampled the image in a task-dependent manner, guiding search towards the target. Human visual search demonstrated invariance in being able to locate objects that were transformed between the target image and the search image in terms of their size (Experiments 1, 2, 3), 2D rotation (Experiments 1 and 2), 3D rotation (Experiment 2), color (Experiment 3), different exemplars from the same category (Experiments 1 and 2), and other appearance changes including occlusion (Experiments 2 and 3). The large dissimilarity between how the targets were rendered in the search image and their appearance in the target image indicates that humans do not merely apply pixel-level template matching to find the target. These results suggest that the features guiding visual search must be invariant to target object transformations.

There has been extensive work characterizing the features that guide visual search[9]. The proposed IVSN model incorporates those ideas into a quantitative image-computable framework to explain how the brain decides where to allocate attention in a task-dependent manner. The IVSN model is agnostic as to whether those attention changes are manifested through overt attention (moving the eyes) or covertly (without moving the eyes). Covert attention changes are harder to quantify at the behavioral level. In the experiments presented here, subjects were instructed to move their eyes to find the target as rapidly as possible. No feedback was provided during the experiment and no punishment was introduced for active exploration via eye movements. The objective was to encourage natural visual search behavior, and avoid alternative strategies such as fixating on the center, and covertly shifting attention until the target was located. The average eye movement reaction times were quite fast (**Fig. S2**) and were consistent with previous work (e.g.,[10]). While we cannot exclude the possibility that there were covert attention shifts in between saccades, there are only a few tens of milliseconds between the first saccade times (**Figures 3D**, **4D**, **5D**) and the latencies that characterize the visually selective responses along the ventral visual cortex (e.g.,[44]), which does not leave much time for extensive processing or multiple attention shifts.

A large body of visual search studies has focused on finding identical matches to a target (e.g.,[9,10,15]). Visual search in the natural world, and most applications of visual search, rarely have the luxury of dealing with identical target search. As expected, performance in such target-identical trials is better than in trials where the target changes shape, both for human subjects as well as for the IVSN model (**Fig. S3**, **S9C**). Furthermore, even the structure and instructions in the task can have an impact on the results. For example, subjects showed higher performance when all the target-identical trials were blocked (**Fig. S9D**). Enhanced performance in blocked identical trials may explain why the overall performance in Experiment 1 was slightly lower than in the study of reference[10]. Of course, there are no blocks of target-identical trials in real world visual search and therefore the mixed conditions of Experiments 1-3 better reflect natural search behavior. These results emphasize the need to use randomized trials and transformed versions of the target object to study real world visual search.

The problem of identifying objects invariantly to image transformations has been extensively discussed in the visual recognition literature (e.g.,[1,2,4], among many others). Indeed, the ventral visual cortex module in IVSN is taken from a computational model that is successful in object recognition tasks, VGG-16[3]. The invariance properties in IVSN are thus inherited from VGG-16. The current results show that the types of features learned upon training VGG-16 in an independent object labeling task (ImageNet[45]), can be useful not only in a bottom-up fashion for visual recognition, but also in a top-down fashion to guide feature-based attention changes during visual search. The model assumes that the same bottom-up features are used in a top-down fashion during visual search. It remains unclear whether bottom-up synaptic weights are directly matched by top-down synaptic weights in cortex and this assumption will require further evaluation through behavioral and physiological experiments. The current results show that top-down features guiding visual search must show invariance to object transformations.

There is no additional training in IVSN to achieve such invariance. The current model, as well as other models of feature-based attention[10,11,29,33,46], assume that such top-down influences provide feature-selective and transformation-tolerant information, as opposed to top-down signals providing non-specific general arousal or gain-control mechanisms. The lack of any training or fine-tuning in the IVSN model distinguishes the proposed model from other work in the object detection literature that focuses on supervised learning from a large battery of similar examples to locate a target[21,22]. The ability to perform a task without extensive supervised learning by extrapolating knowledge from one domain to a new domain is usually referred to as "zero-shot training". The specific exemplar objects in Experiments 1 and 2 were new to the subjects, even though it is reasonable to assume that subjects had had extensive experience with the categories of objects shown in Experiments 1 and 2. Furthermore, subjects were also able to efficiently search for novel objects from novel categories that they had never encountered before (**Fig. S10**). The IVSN model was able to find novel objects from known categories in Experiment 1. More strikingly, the IVSN model could find target objects in natural images even when those objects came from categories that it had never encountered before (Experiment 2, **Fig. S5**). Furthermore, the IVSN model could find Waldo in images that did not resemble any of the images used to train VGG-16 (Experiment 3). The ability to generalize and search for novel objects that have never been encountered before is consistent with the psychophysics literature showing that there are common feature attributes that guide visual search[9]. The proposed IVSN model extends and formalizes the set of attributes from the low-level features that have been extensively studied in psychophysics experiments (e.g. color, orientation, etc.) to a richer and wider set of transformation-tolerant features that have been shown to be useful for visual recognition and which can be used for general visual search under natural conditions.

Beyond exploring average overall performance, it is interesting to examine performance in individual trials and the details of the spatiotemporal sequence of fixations for individual images. There is a large degree of variability when scrutinizing visual search at this high-resolution level. The same subject may follow a somewhat different eye movement trajectory when presented with the same exact target image and search image (**Figure 6**, **Fig. S7**, Fig. **S8**), an effect that cannot be accounted for by memory for the target locations (**Fig. S7**). As expected, the degree of self-consistency was higher than the degree of between-subject consistency, which was in turn higher than the degree of subject model consistency at the image-by-image level both in terms of the number of fixations (**Fig. S7**) as well as in terms of the spatiotemporal sequences of fixations (**Figure 6**, **Fig. S8**).

The current study focuses on the generation of an adequate attention map to guide visual search. Even when IVSN may approximate average human search behavior under some conditions, the model may not be searching in the same way that humans do. There are several important components of visual search that were simplified in the current model but play an important role in real world visual search, and which may contribute to the enhanced between-subject consistency compared to model-subject consistency in individual images. First, the current implementation of the model shows constant acuity over the entire visual field, which is clearly not the case for human vision where acuity drops rapidly from the fovea to the periphery. Combined with potential distance-dependent costs for making saccades (**Fig. S11G-I**), such eccentricity-dependent acuity may play an important role in biasing the attention map and hence directing saccades. Second, once a saccade is made, it is important to decide whether the target is present or not. In the default IVSN model, we did not model this recognition component of visual search here; instead, we used an "oracle" system that perfectly decided whether the fixation window was within the ground truth location of the target object or not (the same oracle detection definition was used throughout for the human psychophysics data for fair comparison, except in **Fig. S12**). As a toy proof-of-principle demonstration, we implemented a simple recognition step for each fixation in **Fig. S11A-C**. As expected, the IVSN$_{recognition}$ model with this recognition module performed very well in Experiment 1, but slightly less well in Experiments 2 and 3 where there is significant clutter. There has been extensive work on invariant visual recognition systems that could be incorporated into IVSN to decide whether the target is present or not[1,3,5,43]. Humans also make recognition mistakes (e.g., there are several examples in **Figures 4C** and **5C** where subjects moved their eyes to the correct location yet did not click the mouse to indicate that they had found the target, see also **Fig. S12**). Third, humans also revisit the same location even if the target is not there (e.g., **Figure 4C**, **5C**, **3E**, **4E**, **5E**,[47,48]). Yet, the default IVSN model implements infinite inhibition of return as a simplifying assumption that could also be improved upon by including a memory decay function, as shown in in the $IVSN_{fIOR}$ model in **Fig. S11D-F**. Fourth, there is no learning in the current model. The visual system needs to be able to learn how to generate a sequence of fixations, including the interaction of the different bottom-up, top-down, memory and recognition components. An elegant idea on how learning could be implemented was presented in ref.[39] where the authors proposed an architecture that can learn to generate eye movements via reinforcement learning with a system that is rewarded when the target is found. The generation of the attention map in the IVSN model is end-to-end trainable. IVSN can be improved by training or fine-tuning via reinforcement learning for various search tasks depending on the applications. Fifth, the images in Experiment 3, and particularly those in Experiment 1, violate basic components of real world images. In real world images, subjects may capitalize on high-level knowledge about scenes[9,49] including understanding certain statistical correlations in object positions (e.g., it is highly unlikely that the car keys would be glued to the ceiling), basic properties of the physical world (an object needs support and therefore keys are more likely to be found on top a desk or the floor rather than floating in the air), correlations in object sizes (the size of a phone in the image may set an expectation for the size of the keys), etc. Such knowledge can place significant constraints on the visual search problem, leading to adequately skipping search over large parts of an image. None of this high-level knowledge is incorporated into the IVSN model.

As emphasized in the previous paragraph, there are multiple directions to improve our quantitative understanding of how humans actively explore a natural image during visual search. The current model provides a reasonable initial sketch that captures how humans can *selectively* localize a target object amongst distractors, the *efficiency* of visual search behavior, the critical ability to search for an object in an *invariant* manner, and zero-shot *generalization* to novel objects including the famous Waldo. Waldo cannot hide any more.

**Materials and Methods**

**Psychophysics experiments**

*Participants*. We conducted four psychophysics experiments with 60 naive observers (19-37 years old, 35 females, 15 subjects per experiment). We focus on the first 3 experiments in the main text and report the results of the fourth experiment in **Fig. S10**. All participants had normal or corrected-to-normal vision. Participants provided written informed consent and received 15 USD per hour for participation in the experiments, which typically took an hour and a half to complete. All the psychophysics experiments were conducted with the subjects' informed consent and according to the protocols approved by the Institutional Review Board.

*Experimental protocol*. The general structure for all three experiments was similar (**Figure 1**). Subjects had to fixate on a cross shown in the middle of the screen, a target object was presented followed by another fixation delay (Experiments 1 and 2), a search image was presented, and subjects had to move their eyes to find the target. In Experiments 2 and 3, subjects also had to indicate the target location via a mouse click. Stimulus presentation was controlled by custom code written in MATLAB using Version 3.0 of the Psychophysics Toolbox[50]. Images were presented on a 19-inch CRT monitor (Sony Multiscan G520), at a 1024x1280 pixel resolution, subtending approximately 32x40 degrees of visual angle. Observers were seated at a viewing distance of approximately 52 cm. We recorded the participants' eye movements using the EyeLink D1000 system (SR Research, Canada).

*Experiment 1 (Object arrays)*. We selected segmented objects without occlusion from 6 categories in the MSCOCO dataset of natural images[40]: sheep, cattle, cats, horses, teddy bears and kites (e.g., **Figure 3A**). Due to the uncontrolled and diverse nature of stimuli in the MSCOCO dataset, the images may differ in low-level properties that could contribute to visual search performance. To minimize such contributions, we took the following steps: (1) resized the object areas such that a bounding box of 156 x 156 pixels encompassed the outermost contour of the object while maintaining their aspect ratios; (2) converted the images to grayscale; (3) equalized their luminance histograms, and (4) randomly rotated the objects in 2D. We conducted a verification test to make sure that the low-level features of all the objects were minimally discriminative: we considered the feature maps from the first convolution blocks of four pre-trained image classification networks (ResNet[43], AlexNet[5], VGG16 and VGG19[3]), and performed cross-validated category classification tests on these features maps as well as on the image pixels using a Support Vector Machine (SVM) classifier [51]. The total of 2000 object images were split into 5 groups for training, validation and testing. The classification performance obtained with these low-level features was consistent across the different computational models and was slightly above chance levels (**Table S1**).

A schematic of the sequence of events during the task is shown in **Figure 1A**. After fixation for 500 ms, a random exemplar from the target category was shown in the fixation location, subtending 5.5 degrees of visual angle, for 1500 ms. The object was shown at a random rotation (0-360 degrees) along with the category name. After another 500 ms of fixation, the search image was presented. Subjects searched for the target in a search image containing an array of 6 objects (**Figure 3A**). In the search images, the 6 objects, each 156 x156 pixels and subtending ~5 degrees of visual angle, were uniformly distributed on a circle with a radius of 10.5 degrees eccentricity. All the objects could be readily recognized by humans at this size and eccentricity. The target was always present only once within these 6 objects and was placed randomly in one of the 6 possible positions. **Fig. S1A** shows the distribution of target object locations. There was one distractor from each category, randomly chosen.

Subjects were instructed to find the target as soon as possible by moving their eyes and pressed a key to go to the next trial. To evaluate within-subject consistency, and unbeknown to the subjects, each trial was shown twice (the exact same target image and search image was repeated). The order of all trials was randomized. There were 300x2=600 trials in total, divided into 10 blocks of 60 trials each. We split the 300 unique trials into 180 target-different trials and 120 target-identical trials (**Fig. S9A**). In the target-identical trials, the appearance of the target object within the search image was identical to that in the target image. In the target-different trials, the target object was a random exemplar from the same category as the one shown in the target image, and was presented at a random rotation (0-360 degrees). Target-different and target-identical trials were randomly interleaved, except in the additional experiment discussed in **Fig. S9D** (see below). To evaluate between-subject consistency, the same target and search images were shown to different subjects.

We initially hypothesized that performance would be higher in target-identical trials compared to target-different trials. Upon examining the results, this hypothesis was found to be correct but the difference in performance between target-identical and target-different trials was small (**Fig. S9C**). In addition, performance in the target-identical trials was lower than what we reported previously in a different experiment consisting exclusively of target-identical trials and using different objects[10]. We conjectured that the task instructions and structure including the presence of target-different trials influenced performance in the target-identical trials. To further investigate this possibility, we conducted an additional variation of Experiment 1 in which target-identical and target-different trials were blocked (**Fig. S9D**). In this task variation, subjects were told whether the next block would include target-identical or target-different trials. To counter-balance any presentation order biases, we tested 2 subjects on target-identical trials first followed by target-different trials and 3 subjects on the reversed order. This experiment confirmed our intuitions and showed that performance was higher in target-identical trials when they were blocked, compared to when they were interleaved, while performance in target-different trials did not depend on the task structure and instructions. Throughout the text (and except for **Fig. S9D**), we focus all the analyses on the original and more natural version of the task where target-identical and target-different trials were randomly interleaved.

*Experiment 2 (Natural images)*. We considered 240 objects from common object categories, such as animals (e.g., clownfish) and daily objects (e.g., alarm clock). The object sizes were 106.5±71.9 pixels high x 114.4±74.8 pixels wide. The 240 objects were *not* restricted to the 6 categories in Experiment 1 but could involve any object. To test whether IVSN can generalize to searching for novel objects (zero-shot training), we also included objects that are *not* part of the 2012 ImageNet data set[45] (the database of images used to train the model, see Model section below). Examples of such objects include SpongeBob toys, Eve robot, Ironman figures, QuickTime app icon, deformed flags or clothes, weapons, tamarind fruits, fried chicken wings, special hand gesture, Lego blocks, push toys, chopsticks, and ribbons on gifts, among others. There were 140 images out of the selected 240 images containing target objects that were not included in ImageNet. All target objects were manually selected such that each search image contained only one target object. The object shown in the target image was *not* segmented from the search image, but rather was a similar object: for example, **Figure 4A** shows a vertically and rotated version of "Minnie" with a dress and bow displaying white circles (left) whereas the target as rendered in the search image shows Minnie at a different scale, with a different attire, partially occluded and under different rotation (right). The search images were 1028 x1280 pixel natural images that contained the target amidst multiple distractors and clutter (e.g., **Figure 4A**). Both the search images and the target images were presented in grayscale. As illustrated in **Figure 4A**, the target objects were picked such that they were visually different from the ones rendered on the search images; these changes included changes in scale, 2D and 3D rotation, changes in attire, partial occlusion, etc.

The sequence of steps in Experiment 2 followed the one described for Experiment 1 (**Figure 1B**), with three differences described next. The presentation of the target image did not include any text. The search image was a grayscale natural image, always containing the target, and occupied the full monitor screen (subtending ~32x40 degrees of visual angle). **Figure S1B** shows the distribution of target object sizes and locations within the search image, which were approximately uniformly distributed. The appearance of the target object within the search array was always different from that in the target image, that is, there were no target-identical trials. Subjects were instructed to find the target as soon as possible by moving their eyes. Experiment 2 was harder than Experiment 1 because objects in the search image were not segmented and were shown embedded in complex natural clutter, and because the appearance of the target object was more different from the target object than in Experiment 1 (e.g. compare x-axis in **Fig. S3A** versus **S3C** versus **S3E**). As the search task became more difficult, subjects could fixate on the target object, yet fail to realize that they had landed on the target (**Fig. S12**). Hence, to ensure that subjects had consciously found the target, they had to use the computer mouse to click on the target location. If the clicked location fell within the ground truth, subjects went on to the next trial; otherwise, subjects stayed on the same search image until the target was found. If the subjects could not find the target within 20 seconds, the trial was aborted, and the next trial was presented. Subjects were unable to find the target within 20 seconds in 16.4% of the trials. To evaluate between-subject consistency, different subjects were presented with the same images. To evaluate within-subject consistency, every trial was repeated once, in random order (same target image and same search image). To avoid any potential memory effect (whereby subjects could remember the location of the target), we restricted the analyses to the first presentation, except in the within-subject consistency metrics reported in **Figure 6**, **Fig. S7** and **S8**. The results were very similar for the first instance of each image versus the second instance of each image and any memory effects across trials were minimal, but we still implemented these precautions focusing the results on the first instance of each image in all the experiments.

*Experiment 3 (Waldo images)*. 'Where's Waldo' is a well-known search task[41] with crowded scene drawings containing hundreds of individuals that look similar to Waldo undertaking various activities. Exactly one of these individuals is the character known as Waldo (e.g., **Figure 5A**). We tested 67 Waldo images from ref[41]. The target object sizes were 24.7±4.5 pixels wide and 40.3x7.4 pixels high. Given the large size of the Waldo search images and the limited precision of our eye tracker in terms of individual characters on these images, we cropped each Waldo image into four quadrants and only showed the human subjects the quadrant containing Waldo. There were 13 out of 67 images that had an instruction panel in the upper left corner that could contain additional renderings of Waldo. Subjects were explicitly instructed not to look at the instruction panel. At the model evaluation stage, these areas were also discarded. The locations of these panels can be approximately glimpsed from less dense fixation patches in **Fig. S1H.** Because all subjects were familiar with the Waldo task, we changed the overall structure such that there was no target image presentation in each trial (**Figure 1C**). The target (Waldo) in color was presented at the beginning of the experiment. After fixation, the search image, always containing Waldo, was presented occupying the full monitor screen (subtending ~32x40 degrees of visual angle). Subjects were instructed to find Waldo as soon as possible by moving their eyes. Similar to Experiment 2, once the target was found, subjects had to click on the target location. If the clicked location fell on the ground truth, subjects proceeded to the next trial; otherwise, subjects stayed on the same search image until the target was found. If subjects could not find the target in 20 seconds, the trial was aborted. The limit of 20 seconds was based on pilot tests and was dictated by a compromise between allowing enough time to find the target in as many trials as possible while at the same time maximizing the number of search trials. Subjects were unable to find the target within 20 seconds in 27% of the trials. There were 67 trials in total and the trial order was randomized. Within- and between-subject consistency was evaluated as described above for Experiments 1 and 2. In addition to searching for Waldo, we conducted a separate set of trials where subjects searched for the

'Wizard', another character in the Waldo series. The results for the Wizard search were similar to those for the Waldo search. We restrict this report to the Waldo search task for simplicity.

*Experiment 4 (Novel objects)*. We conducted an additional experiment to evaluate whether human subjects are able to search for novel objects that they have never encountered before (other than the single exposure to the target image). We collected a total of 1860 novel objects belonging to 98 categories. These objects were composed from well-designed novel object parts and we also included novel objects used in previous studies (**Fig. S10**)[52,53]. We used the same pre-processing steps to normalize the novel objects' low-level features as in Experiment 1. F**ig. S10A** shows 6 example novel objects. The task structure followed the one in Experiment 1, except that here there was no text indicating the object category during the target presentation (**Fig. S10B**). The number of trials for target identical and target different trials was balanced (80 target-identical vs. 80 target-different trials in novel objects). To directly compare the results for novel objects versus those obtained with known objects, the objects from Experiment 1 (known objects) were also presented in this experiment, randomly intermixed with the novel object trials.

In visual search experiments, the similarity between the target object and the distractor objects plays a critical role in the difficulty of the task. As a proxy for task difficulty, we computed the similarity between the target object and the distractors by computing the Euclidian distance between all possible target-distractor object pairs in each image (x-axis in **Fig. S10C**). The target and distractor novel objects were chosen so as to match the distribution of similarities for known objects (**Fig. S10C**) to avoid scenarios where one set of stimuli could be easier to discriminate than in the other set. The results for the novel object visual search experiment are shown in **Fig. S10D** and **S10E**.

**Visual search computational models**

We first provide a high-level intuitive outline of our invariant visual search network (IVSN) model, followed by a full description of the implementation details. IVSN posits an attention map, $M_f$, which determines the fixation location by conjugating local visual inputs with target information (**Figure 2**). Both the target image ($I_t$) and the search image ($I_s$) are processed through the same deep convolutional neural network, which aims to mimic the transformation of pixel-like inputs through the ventral visual cortex[1,2,4]. Feature information from the top level of the visual hierarchy is stored in a module which we refer to as pre-frontal cortex, based on the neurophysiological role of this area during visual search (e.g.,[15]). Activity from the pre-frontal cortex module provides top-down modulation, based on the target high-level features, on the responses to the search image, generating the attention map $M_f$. A winner-take-all mechanism selects the maximum local activity in the attention map $M_f$ for the next fixation. If the fixation location contains the target, the search stops. Otherwise, an inhibition-of-return mechanism leads the model to select the next maximum in the attention map and the process thus continues until the target object is found. The model was always presented with the exact same images that were shown to the subjects in the psychophysics experiments described in the previous section.

*Ventral visual cortex*. The deep feed-forward network builds upon the basic bottom-up architecture for visual recognition described in previous studies (e.g.,[1-8]). We used a state-of-the-art deep feed-forward network, implemented in VGG16[3], pre-trained for image classification on the 2012 version of the ImageNet dataset[45]. The network weights **W** learnt from image classification extract feature maps for an input image of size 224 x224 pixels. The same set of weights, that is, the same network, is used to process the target image and the search image. Only a subset of the multiple layers is illustrated in **Figure 2** for simplicity (see ref.[3] for full details of the VGG16 architecture). The images from the ImageNet dataset used to train the ventral visual cortex network for object classification are different from *all* the images used in the experiments. The 6 categories from MSCOCO in Experiment 1 are also present in ImageNet. In Experiment 2, 140 of the 240 target objects were *not* part of the 1000 ImageNet categories. None of the images in Experiment 3 or in the novel object experiment (**Fig. S10**) had any resemblance to the categories in ImageNet. The weights **W** do *not* depend on any of the target images $I_t$ or the search images $I_s$ (hence the model constitutes a zero-shot training architecture for visual search). The output of the ventral visual cortex module is given by the activations at the top-level (Layer 31 in VGG16[3]), $\varphi_{31}$ *($I_t$, **W**)*, and the layer before that (Layer 30 in VGG16), $\varphi_{30}$ *($I_s$, **W**)*, in response to the target image and search image, respectively (in **Fig. S13** we considered top-down modulation between different layers). As noted above, it is the same exact network, with the same weights **W** that processes the target and search images, and we use the activations in layer 31 in response to the target image to provide top-down modulation to layer 30's response to the search image (**Figure 2**). In Experiments 2-3, the images were too large (1080 x 1240 pixels) for the model and down-sampling the images would make the finely detailed characters hard to discern. Therefore, we partitioned the whole image into segments of size 224 x 224, repeatedly ran the model in each of these segments and finally concatenated the resulting attention maps.

*Pre-frontal cortex*. The top-level of the VGG-16 architecture conveys the target image information to the pre-frontal cortex module, consisting of a vector of size 512. To search for the target object, IVSN uses the ventral visual cortex responses to that target image stored in the pre-frontal cortex to modulate the ventral visual cortex responses to the search image. This modulation is achieved by convolving the representation of the target with the representation of the search image before max-pooling:

$$M_f = m(\varphi(I_t, W), \varphi(I_s, W)) = m(\varphi_{31}(I_t, W), \varphi_{30}(I_s, W))$$

where *m(.)* is the target modulation function defined as a 2D convolution operation with kernel $\varphi_{31}$ *($I_t$, **W**)* on the search feature map $\varphi_{30}$ *($I_s$, **W**)*. $M_f$ denotes the attention map.

*Fixation sequence generation*. At any point, the maximum in the attention map determines the location of the next fixation. In the figures, we normalize the attention map to [0,1] for visualization purposes.

A winner-take-all mechanism selects the fixation location. The model needs to decide whether the target is present at the selected location or not (see below). If the target is located, search ends. Otherwise, inhibition-of-return[48] is applied to $M_f$ by reducing the activation to zero in an area of pre-defined size (45x45 pixels in Experiment 1, 200x200 in Experiment 2, 100x100 in Experiment 3), centered on the current fixation location. This reduction is permanent, in other words, infinite memory is assumed for inhibition of return here. These window size choices were based on the average object sizes in each experiment. Similar to other attention models (e.g., ref.[26]), the winner-take-all mechanism then selects the next fixation location and this procedure is iterated until the target is found. In the psychophysics experiments, we limited the duration of each trial to 20 seconds. When we compared the number of fixations at the image-by-image level (**Fig. S7**), we restricted the analyses to those images when the target was found and excluded those images where the target was not found in 20 seconds (see previous section for percentages in each task). Otherwise, all images were included in the analyses.

*Target presence decision*. Given a fixation location, the model needs to perform visual recognition to decide whether the target is present or not (in a similar way that humans need to decide whether they found the target after moving their eyes to a new location). There has been extensive work on visual recognition models (e.g., [1,3-5]). In this study, we focus on the attention selection mechanism. To isolate the search process from the verification process, in the default IVSN model we bypass the recognition question by using an 'oracle' system that decides whether the target is present or not (see **Fig. S11A-C** for IVSN$_{recognition}$). The oracle checks whether the selected fixation falls within the ground truth location, defined as the bounding box of the target object. The bounding box is defined as the smallest square encompassing all pixels of the object. For fair comparison between models and humans, we implemented the same oracle system for the human psychophysics data (except in **Fig. S12**), by considering the target to be found the first time a subject fixated on it.

*Comparison with other models*. We performed several comparisons with other models (**Fig. S4**, **S11**, **S13**, **S14**). In all cases, the alternative models proposed a series of fixations. In all cases except for $IVSN_{recognition}$ (described below), we used the oracle method to decide whether to stop search or to move on to the next fixation. In all cases except for $IVSN_{fIOR}$ (described below), the models had infinite inhibition of return (IOR), as described above. We considered the following alternative models:

(1) Chance. We considered a model where the location of each fixation was chosen at random. In Experiment 1, we randomly chose one out of the six possible locations, while still respecting infinite IOR. In Experiments 2 and 3, a random location was selected in each fixation, while still respecting IOR; this random process was repeated 100 times. The selected location was the center of a window of the same size used for the recognition model described above. This window was used to determine the presence of the target and also to set IOR.

(2) Sliding Window (SW). We considered a sliding window approach which takes the fixated area (a window of the same size used for the recognition model described above) as inputs, scans the search image from the top left corner with stride 28 pixels, and uses oracle verification to determine target presence. In Experiment 1, the sliding window sequentially moves through the 6 possible objects.

(3) Template Matching. To evaluate whether pixel-level features of the target were sufficient to direct attention, we introduced a pixel-level template-matching model where the attention map was generated by sliding the canonical target of size 28 x28 pixels over the whole search image. Compared with the SW model, the Template Matching model can be thought of as an attention sliding window.

(4) IttiKoch. It is conceivable that in some cases, attention selection could be purely driven by bottom-up saliency effects rather than target-specific top-down attention modulation. We considered a pure bottom-up saliency model that has no information about the target[26].

(5) RanWeight. Instead of using VGG16[3], pre-trained for image classification, we randomly picked weights W from a Gaussian distribution with mean 0 and standard deviation 1000. The network was otherwise identical to IVSN. We ran 30 iterations of this model, each iteration with random selection of weights.

*Variations and extensions of the IVSN model.* We considered several possible extensions and variations of the IVSN model.

IVSN$_{AlexNet}$ (**Fig. S14**). The "ventral visual cortex" module in **Figure 2** was replaced by the AlexNet architecture[5]. The "pre-frontal cortex" module corresponded to layer 8 and sent top-down signals to layer 7.

IVSN$_{ResNet}$ (**Fig. S14**). The "ventral visual cortex" module in **Figure 2** was replaced by the ResNet200 architecture[43]. The "pre-frontal cortex" module corresponded to the output of residual block 8 in the target image and sent top-down signals to residual block 8 in the search image.

IVSN$_{FastRCNN}$ (**Fig. S14**). The "ventral visual cortex" module in **Figure 2** was replaced by the FastRCNN architecture[21] pre-trained on ImageNet for region proposal and pre-trained on PASCAL VOC for object detection. The "pre-frontal cortex" module corresponded to layer 24 and sent top-down signals to layer 23.

IVSN$_{24 \to 23}$, IVSN$_{17 \to 16}$, IVSN$_{10 \to 9}$, IVSN$_{5 \to 4}$ (**Fig. S13**). In the IVSN model as presented in **Figure 2** (based on the VGG16 architecture[3]), the "pre-frontal cortex" module corresponded to layer 31 and sent top-down signals to layer 30. We considered several variations using top-down features from different levels of the VGG16 architecture as described by the model sub indices.

IVSN$_{recognition}$ (**Fig. S11A-C**). The IVSN model presented in the main text uses an oracle to determine whether the target was found at a given fixation or not. In the brain, of course, there is no oracle. Each fixation places the new location within the high-resolution fovea, and responses along the ventral visual stream within this region are enhanced via attention modulation[15,29,31]. By emphasizing the selected areas, IVSN allows the ventral pathway to perform fine-grained object recognition. As a schematic proof-of-principle of a model that addresses whether the target was found or not, in **Fig. S11A-C** we implemented an additional step that included recognition after fixation. This recognition machinery involved an object classifier which determined whether the fixated area contained the target or not (IVSN$_{recognition}$). We implemented this step by cropping the search image centered at the fixation location using the same window sizes described for inhibition of return (45x45, 200x200, and 100x100, for Experiments 1, 2, and 3, respectively), and using the object recognition network, VGG16[3], pre-trained on ImageNet[45], to extract the classification vector from the last layer, which emulates responses in inferior temporal cortex with high object selectivity and large receptive fields, for both the target image $I_t$ and the cropped area. The Euclidean distance between activation of this top layer to $I_t$ and the cropped area was computed. If this Euclidian distance was below a threshold of 0.9, the target was deemed to be found and search was stopped. Otherwise, the search continued after applying inhibition-of-return, as described above for the oracle. In this model including a recognition component, failure to locate the target could be due to fixating on the wrong location or fixating on the right location but not realizing that the target was there.

IVSN$_{fIOR}$. The IVSN model assumes infinite inhibition-of-return, that is the model never revisits a given fixation location. In contrast, humans do tend to revisit the same location even if the target is not there. An example of this behavior can be seen in multiple fixations from subject 1 in **Fig. 5C** and also in fixations 3 and 6 in **Fig. S7B2** (the reader may have to zoom in on the figures to appreciate this phenomenon). The finite inhibition of return is a well known phenomenon in the psychophysics literature[42,47,48]. We implemented a variation of the IVSN model with finite inhibition-of-return (IVSN$_{fIOR}$). At each location in the image *(x,y)* and at time *t*, the feature attention map $M_f$ was multiplied by a memory function $M_m$ to generate a new attention map $A_f(x,y)=M_f(x,y)*M_m(x,y,t)$. In the implementation with infinite IOR, $M_m(x,y,t)$ is 0 if the location *(x,y)* was visited previously and 1 otherwise (independently of time *t*). In the IVSN$_{fIOR}$ model, $M_m(x,y,t)$ was fitted to the empirical probability of revisiting a location from the human psychophysics data. The inaccuracy in our eye movement measurements is on the order of 1 degree of visual angle. To be overly cautious, we defined a location as revisited if another fixation landed within 3 degrees of visual angle. None of the parameters in the default IVSN model were trained or fitted to human psychophysics data. In contrast, the function $M_m$ was fitted to the human psychophysics data, separately for each Experiment. To avoid overfitting, we randomly selected 7 out of the 15 subjects to fit $M_m$ and all the comparisons between $IVSN_{fIOR}$ and human psychophysics was based on the remaining 8 subjects.

$IVSN_{size}$. The IVSN model has no constrain on the size of each saccade (e.g. one fixation could be in the upper left corner and the immediate next fixation could be in the lower right corner). In contrast, humans tend to make smaller saccades following a gamma-like distribution (**Fig. S11G-I**). We implemented a variation of the IVSN model where the saccade size was constrained by the empirical distribution of human saccade sizes ($IVSN_{size}$). We defined the attention map as a weighted sum of the feature attention map $M_f$ and a size constraint function $M_{sc}$: $A_f(x,y) = w\ M_f(x,y) + (1-w)\ M_{sc}(x,y)$. The weight factor $w$ was set to 0.2346 across all the experiments, selected to optimize the fit between human and $IVSN_{size}$ saccade sizes. In a similar fashion to $IVSN_{fIOR}$ and to avoid overfitting with did cross-validation by fitting $M_{sc}$ separately for each experiment, using only a random subset of 7 out of the 15 subjects.

**Data analysis**

*Psychophysics fixation analysis*. We used the EDF2Mat function provided by the EyeLink software (SR Research, Canada) to automatically extract fixations. We clustered consecutive fixations that were within object bounding boxes of size 45x45 pixels for more than 50ms. If fixation was not detected during the initial fixation window, the experimenter re-calibrated the eye tracker. The last trial before re-calibration and the first trial after calibration were excluded from analyses. In Experiment 1, we filtered out fixations falling outside the six object locations (13.7±5.6% of the trials). Upon presentation of the search image, we considered the first fixation away from the center. We considered that a fixation had landed on the target object if it was within a square window centered on the target object. The window sizes were 45x45 for Experiment 1, 200 x 200 pixels for Experiment 2 and 100 x 100 pixels for Experiment 3. These values correspond to the mean widths and heights of all the ground truth bounding boxes for each dataset (**Fig. S1**). In Experiments 2 and 3, subjects had to click the target location with the mouse. The mouse click location had to fall on the window defining the target object location for the trial to be deemed successful. In 15.9±4.9% of trials in Experiment 2 and 10.1±7.0% of trials in Experiment 3, the initial mouse clicks were incorrect. If the location indicated by the mouse click was incorrect, subjects had to continue searching; otherwise, the trial was terminated. It should be noted that in several cases, subjects could fixate on the target object but not click the mouse, most likely because they were not consciously aware of finding the target despite the correct fixation (**Fig. S12**, see Discussion). As discussed above, for fair comparison with the models, we used an oracle version such that the target was considered to be found upon the first fixation on the target, except in **Fig. S12**.

*Comparisons of fixation patterns*. We evaluated the degree of within-subject consistency by comparing the fixations that subjects made during the first versus second presentation of a given target image and search image. We evaluated the degree of between-subject consistency by performing pairwise comparisons of the fixations that subjects made in response to the same target image and search image for all 15-choose-2 subject pairs. We compared the fixations of the IVSN model against each of the 15 subjects. We used the following metrics to compare fixations within subjects, between subjects and between subjects and the IVSN model: (1) we considered the cumulative accuracy as a function of the number of fixations to evaluate the overall search performance (**Figures 3E**, **4E**, **5E**); (2) we compared the number of fixations required to find the target on an image-by-image basis (**Fig. S7**); (3) we compared the spatiotemporal sequence of fixations on an image-by-image basis (**Figure 6**, **Fig. S8**).

(1) Cumulative performance. We compute the probability distribution *p(n)* that the subject or model finds the target in *n* fixations. **Figures 3E**, **4E** and **5E** show the cumulative distribution of *p(n)*.

(2) Number of fixations to find the target. For each image, we plot the number of fixations required to find the target for S1 and S2 where S1 and S2 can be different repetitions of the same image (within-trial consistency), different subjects (between-trial consistency), or subject and model (model-subject consistency). This metric is reported in **Fig. S7**.

(3) Spatiotemporal dynamics of fixations on an image-by-image basis. We used the scanpath similarity score proposed by Borji et al[27]. This measure takes into account both spatial and sequential order by aligning the scanpath between two sequences. We used the implementation described in reference[54]. Briefly, a mean-shift clustering for all human fixations was computed, and a unique character was assigned to each cluster center and corresponding fixations. The Needleman-Wunsch string match algorithm[55] was implemented to evaluate the similarity of a scanpath pair. In **Fig. S8**, we compare the entire sequences. In **Figure 6**, we compare the first x fixations as shown in the x-axis in the figure.

*Statistical analyses*. We used two-tailed t-tests when comparing two distributions and considered results to be statistically significant when p<0.01. Because calculations of p values tend to be inaccurate when the probabilities are extremely low, we reported all p values less than $10^{-15}$ as $p<10^{-15}$ (as opposed to reporting, for example, $p=10^{-40}$); clearly none of the conclusions depend on this.

**Data Availability**. All the raw data are publicly available through the lab's GitHub repository:
 https://github.com/kreimanlab/VisualSearchZeroShot

**Code availability**. All the source code is publicly available through the lab's GitHub repository:
 https://github.com/kreimanlab/VisualSearchZeroShot

**Acknowledgments**. We thank Farahnaz Wick for comments on the manuscript, and Suresh Krishna and Jeremy Wolfe for discussions.

**Funding**. This work was supported by NIH and NSF grants to GK. This material is based upon work supported by the Center for Minds, Brains and Machines (CBMM), funded by NSF STC award CCF-1231216.

# Figure 1

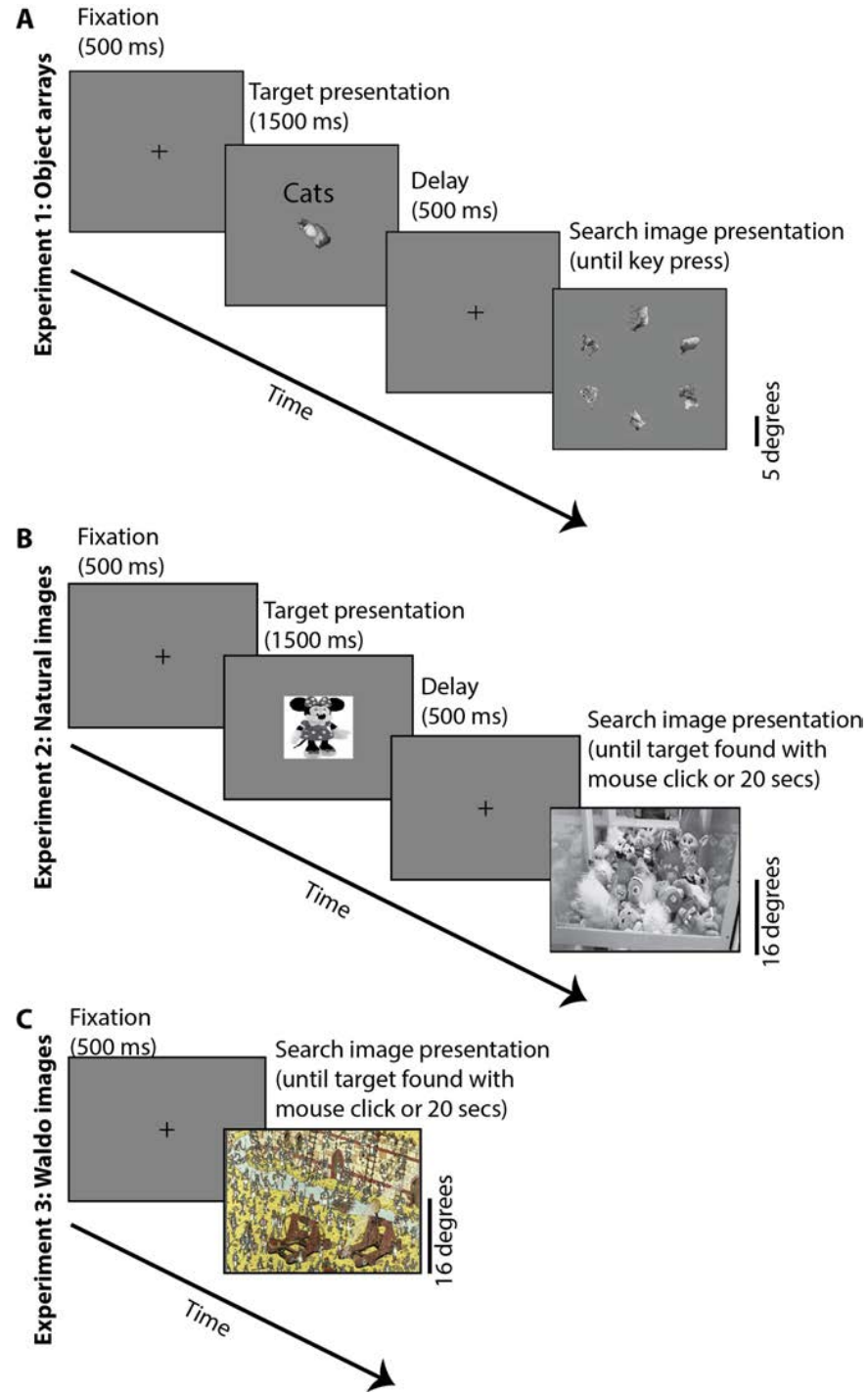

**Figure 1. Schematic description of the three tasks.** **A**. Experiment 1 (Object arrays). **B**. Experiment 2 (Natural images). **C**. Experiment 3 (Waldo images). All tasks started with a 500 ms fixation period. Experiments 1 and 2 were followed by presentation of the target object for 1500 ms. In Experiment 1, the target object appeared at a random 2D rotation and the category descriptor was also shown to emphasize that subjects had to invariantly search for a different exemplar of the corresponding category shown at a different rotation. In Experiment 2, the target object was also different from the rendering in the search image. The target object (Waldo) was not shown in every trial in Experiment 3. In Experiments 1 and 2, there was an additional 500 ms delay after the target object presentation. Finally, the search image was presented and subjects had to move their eyes until they found the target. In Experiments 2 and 3, subjects also had to use the computer mouse to click on the target location.

# Figure 2

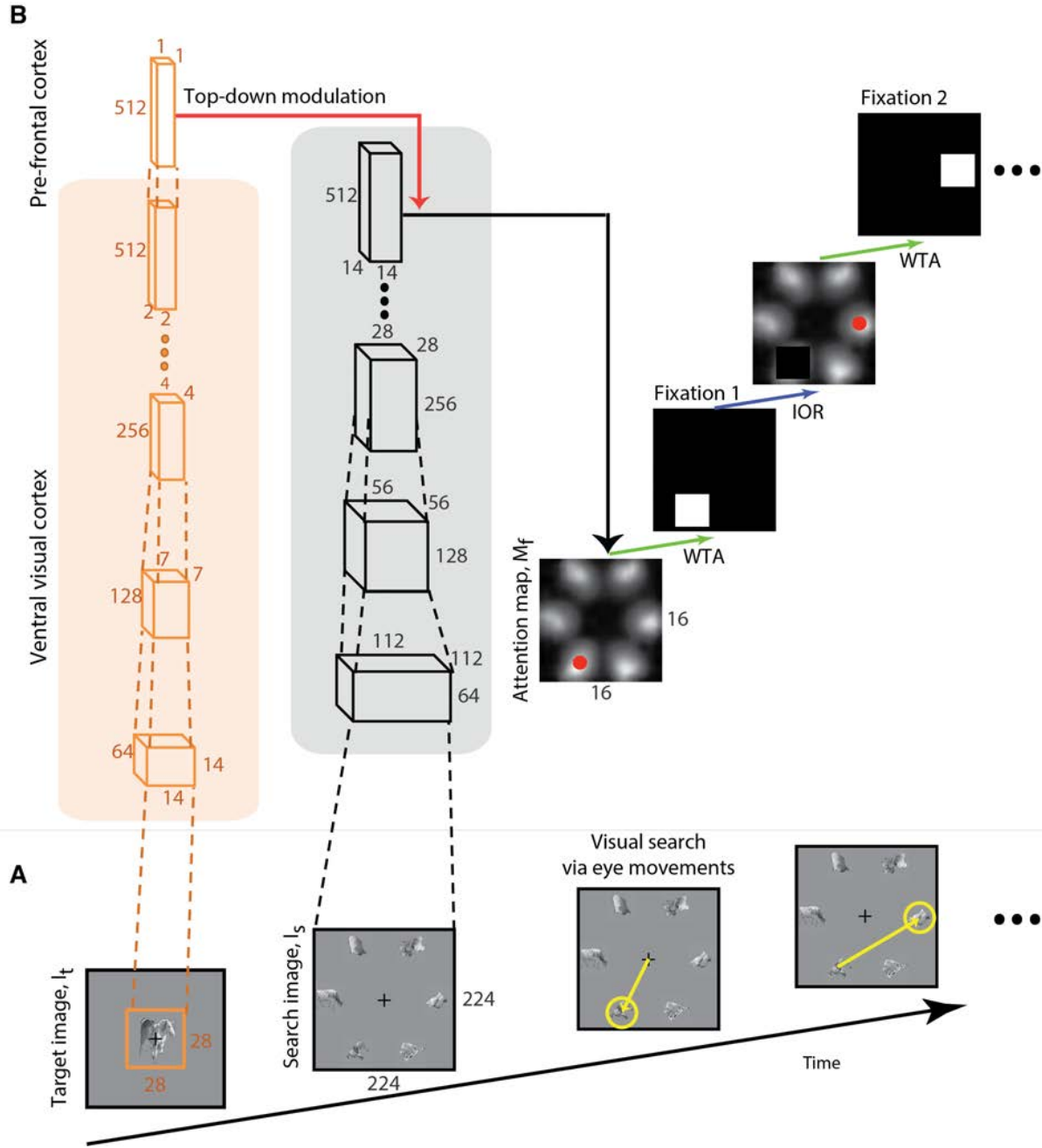

**Figure 2: Model schematic. A.** Sequence of events during the visual search task. A target image is presented, followed by a search image where subjects move their eyes to locate the target object (see **Figure 1** for further details). **B.** Architecture of the model, referred to as Invariant Visual Search Network (IVSN). The model consists of a pre-trained bottom-up hierarchical network (VGG-16) that mimics image processing in the ventral visual cortex for the target image (orange box) and for the search image (gray box). Only some of the layers are shown here for simplicity, the dimensions of the feature maps are indicated for each layer. The model generates features in each layer when presented with the target image $I_t$. The top level features are stored in a pre-frontal cortex module that contains the task-dependent information about the target in each trial. Top-down information from pre-frontal cortex modulates (red arrow) the features obtained in response to the search image, $I_s$, by convolving the target presentation of $I_t$ with the top-level feature map from $I_s$, generating the attention map $M_f$. A winner-take-all mechanism (WTA, green arrow) chooses the maximum in the attention map (red dot) as the location for the next fixation. If the target is not found at the current fixation, inhibition of return (IOR, blue arrow), the fixation location is set to 0 in the attention map and the next maximum is selected. This process is repeated until the target is found.

# Figure 3

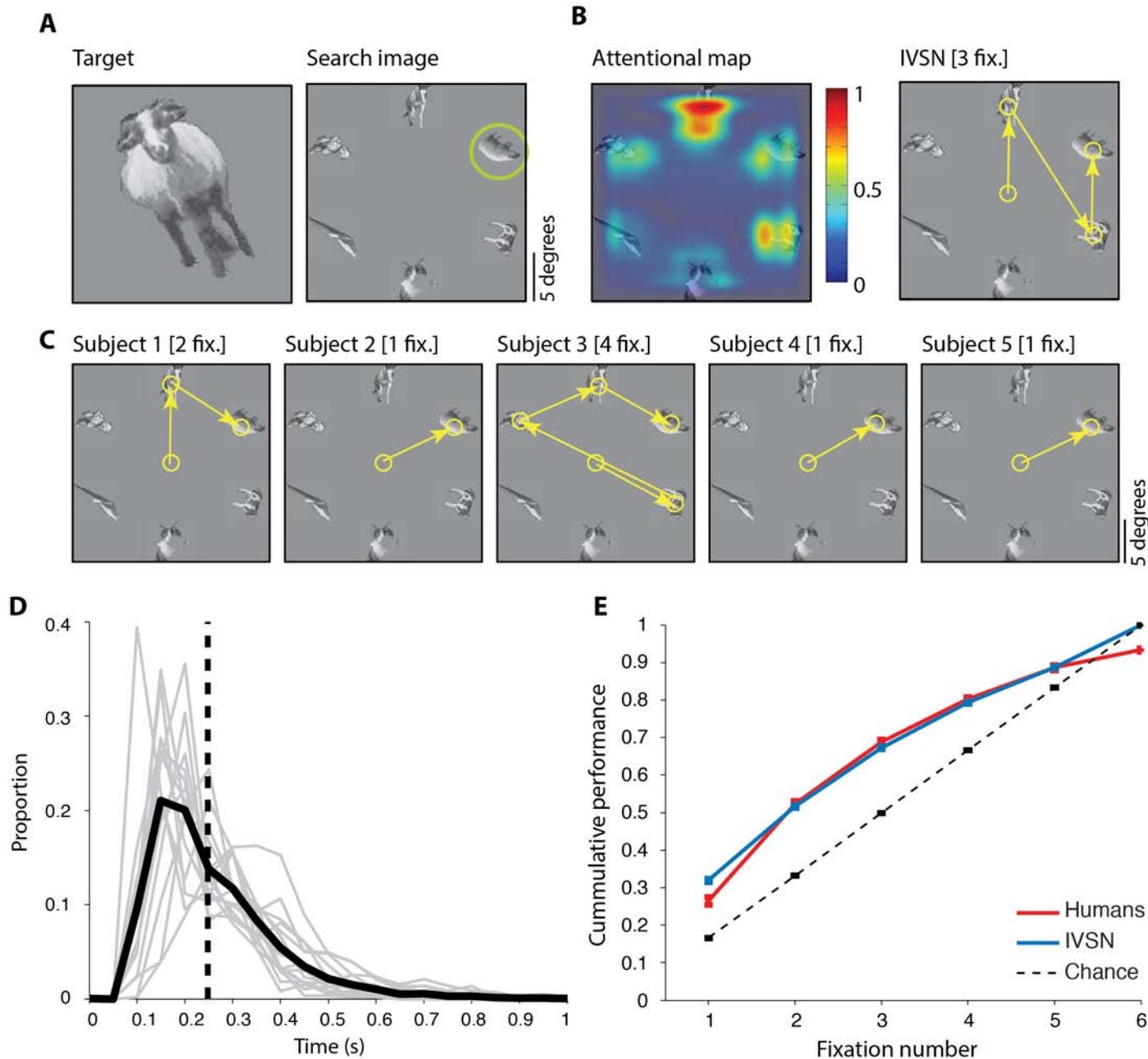

**Figure 3: Experiment 1 (Object arrays).**
**A**. Example target and search images for one trial. The green circle showing the target position was not shown during the actual experiment.
**B**. Attentional map overlaid on the search image and sequence of fixations for the IVSN model.
**C**. Sequence of fixations for 5 subjects (out of 15 subjects). The number above each subplot shows the number of fixations when the target was found.
**D**. Distribution of reaction times for the first fixation (see **Figure S2** for the corresponding distributions for subsequent fixations). Gray lines show the reaction time distributions for 5 individual subjects, the black line shows the average distribution. The median reaction time was 248 ms (SD=130 ms, vertical dashed line).
**E**. Cumulative performance as a function of fixation number for humans (red), IVSN oracle model (blue) and chance model (dashed line). Error bars denote SEM (see **Figure S4** for comparisons against other models).

# Figure 4

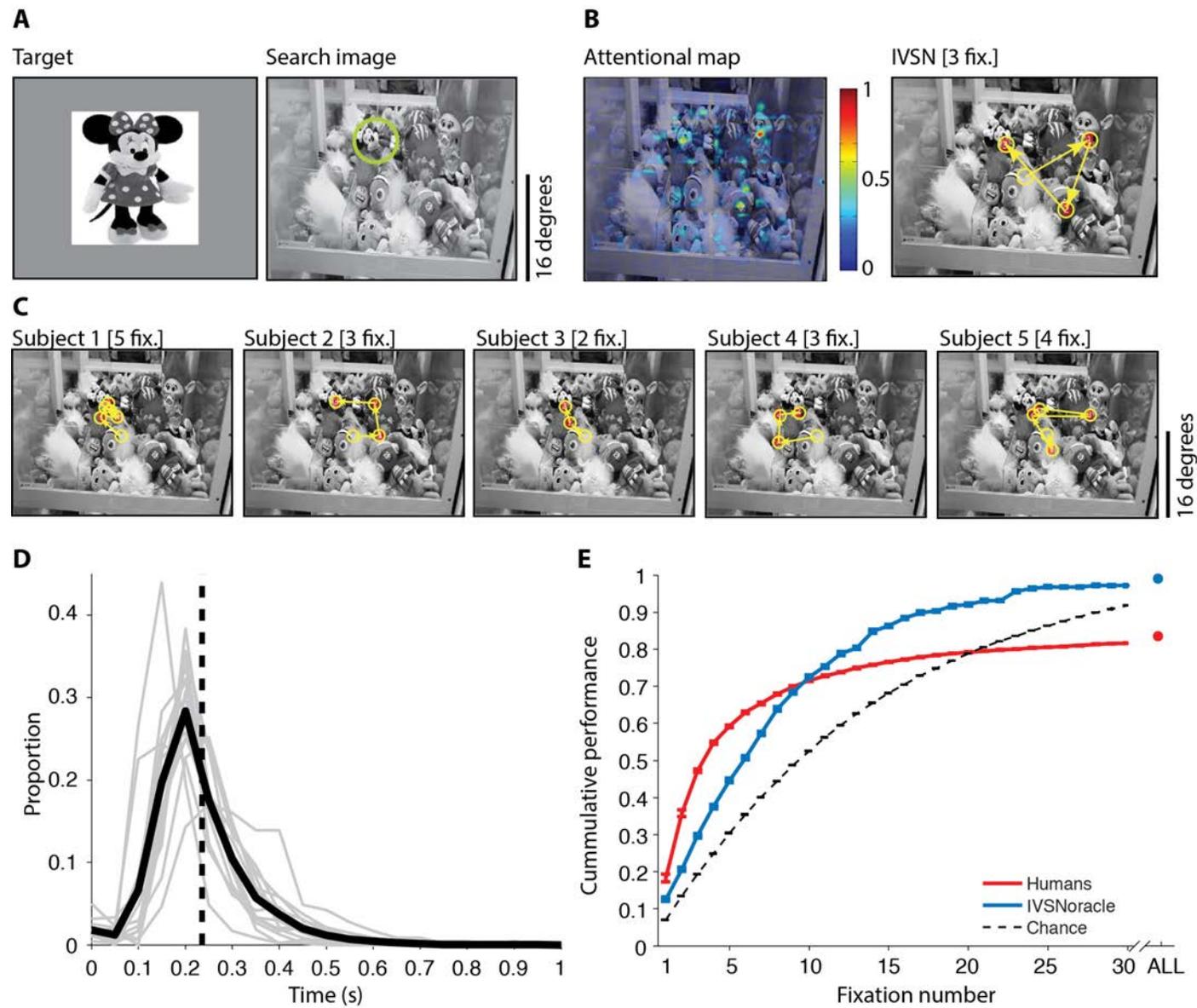

**Figure 4: Experiment 2 (Natural images).** The format and conventions follow those in **Figure 3**.

# Figure 5

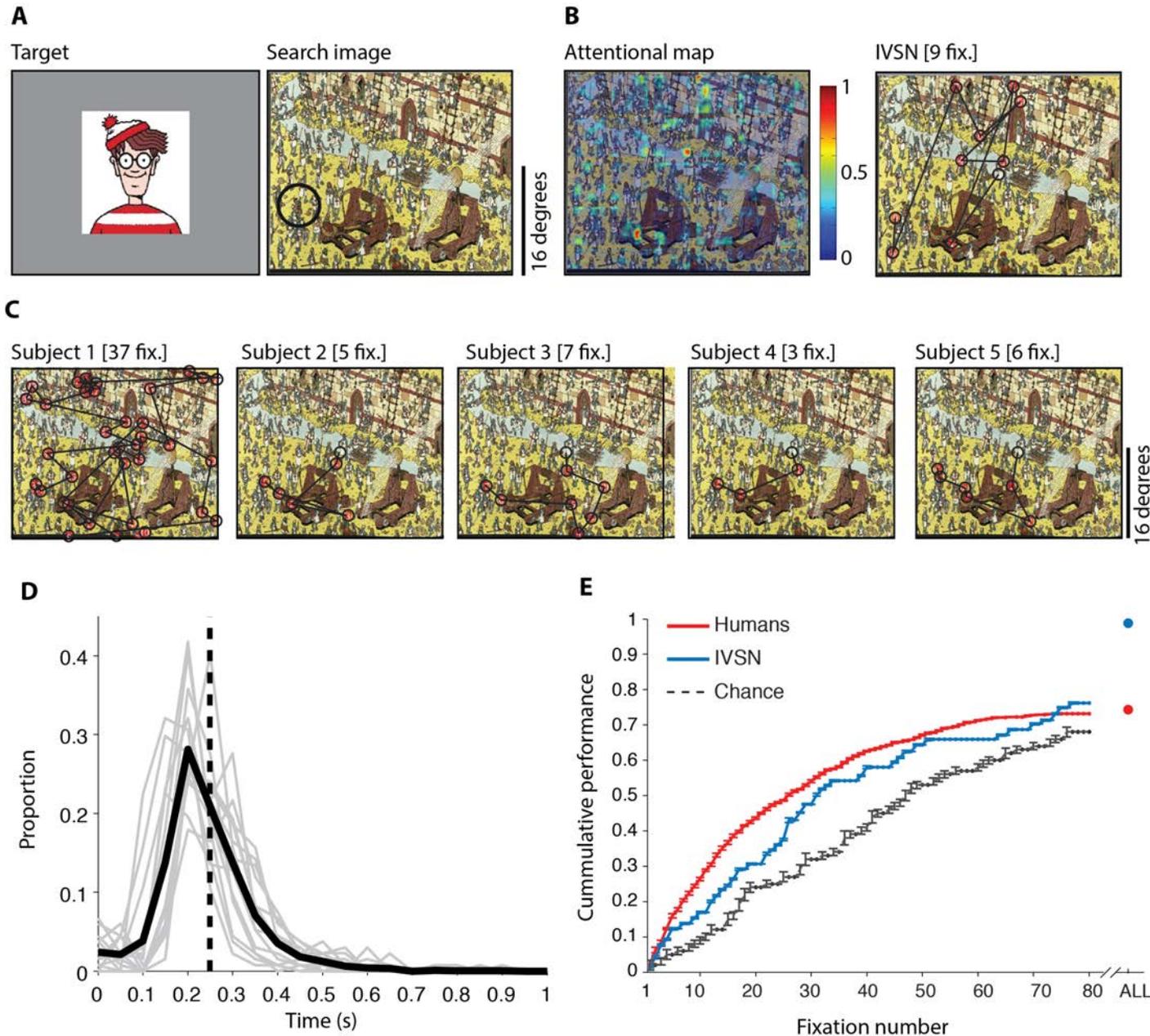

**Figure 5: Experiment 3 (Waldo images)**. The format and conventions follow those in **Figure 3**.

# Figure 6

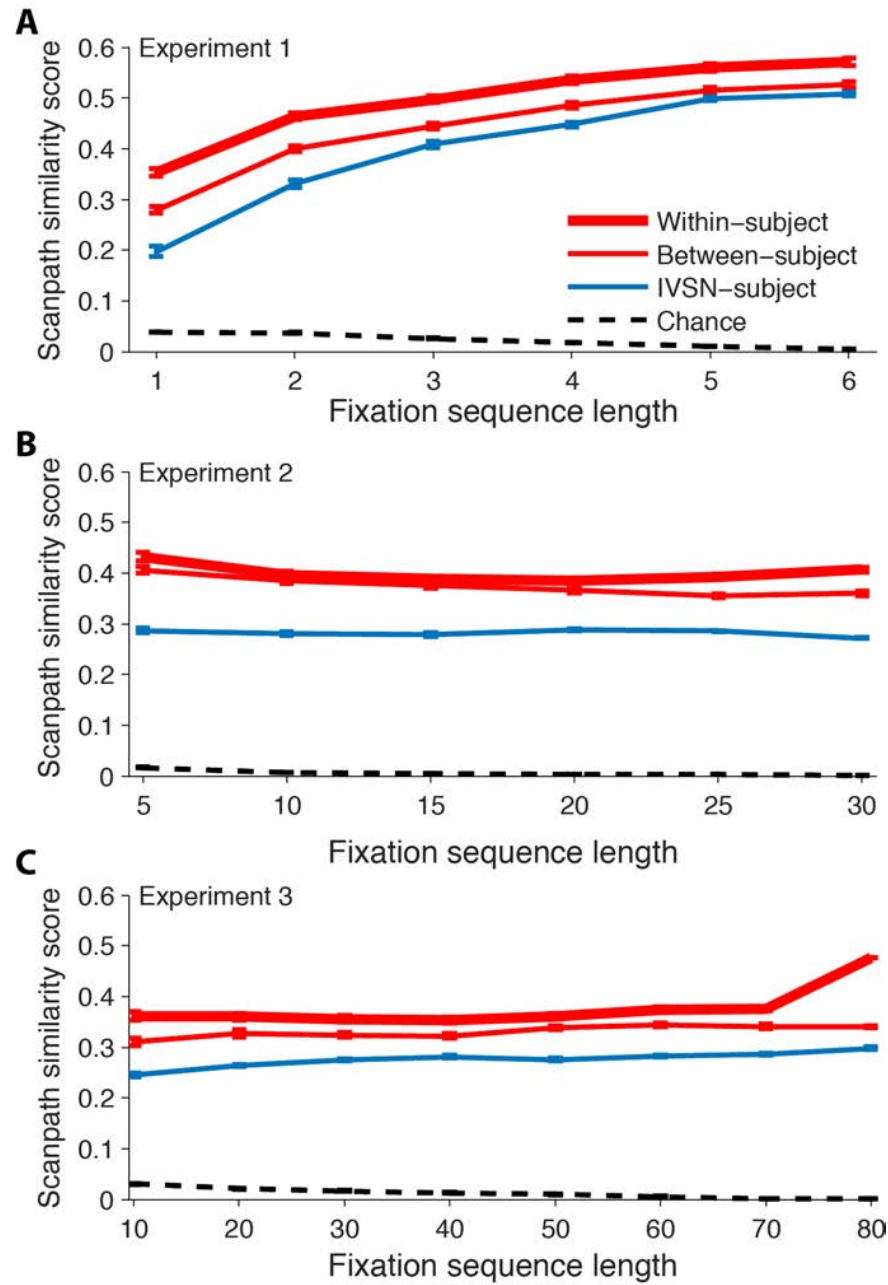

**Figure 6: Image-by-image consistency in the spatiotemporal pattern of fixation sequences.** Scanpath similarity score (see text and **Methods** for definition) comparing the fixation sequences within subjects (thick red), between-subjects (thin red) and between the IVSN model and subjects (blue) for Experiment 1 (**A**), Experiment 2 (**B**), and Experiment 3 (**C**). The larger the scanpath similarity score, the more similar the fixation sequences are. The x-axis indicates the length of sequences compared. For a given fixation sequence length $x$, only sequences of length $\geq x$ were considered and only the first $x$ fixations were considered. The dashed line indicates the similarity between human sequences and random sequences.

The within-subject similarity score was higher than the between-subject score in all 3 experiments ($p<10^{-9}$). The between-subject similarity score was higher than the IVSN-human score in all 3 experiments ($p<10^{-15}$) and the IVSN-human similarity scores were higher than human-chance scores in all 3 experiments ($p<10^{-15}$).

# Figure S1

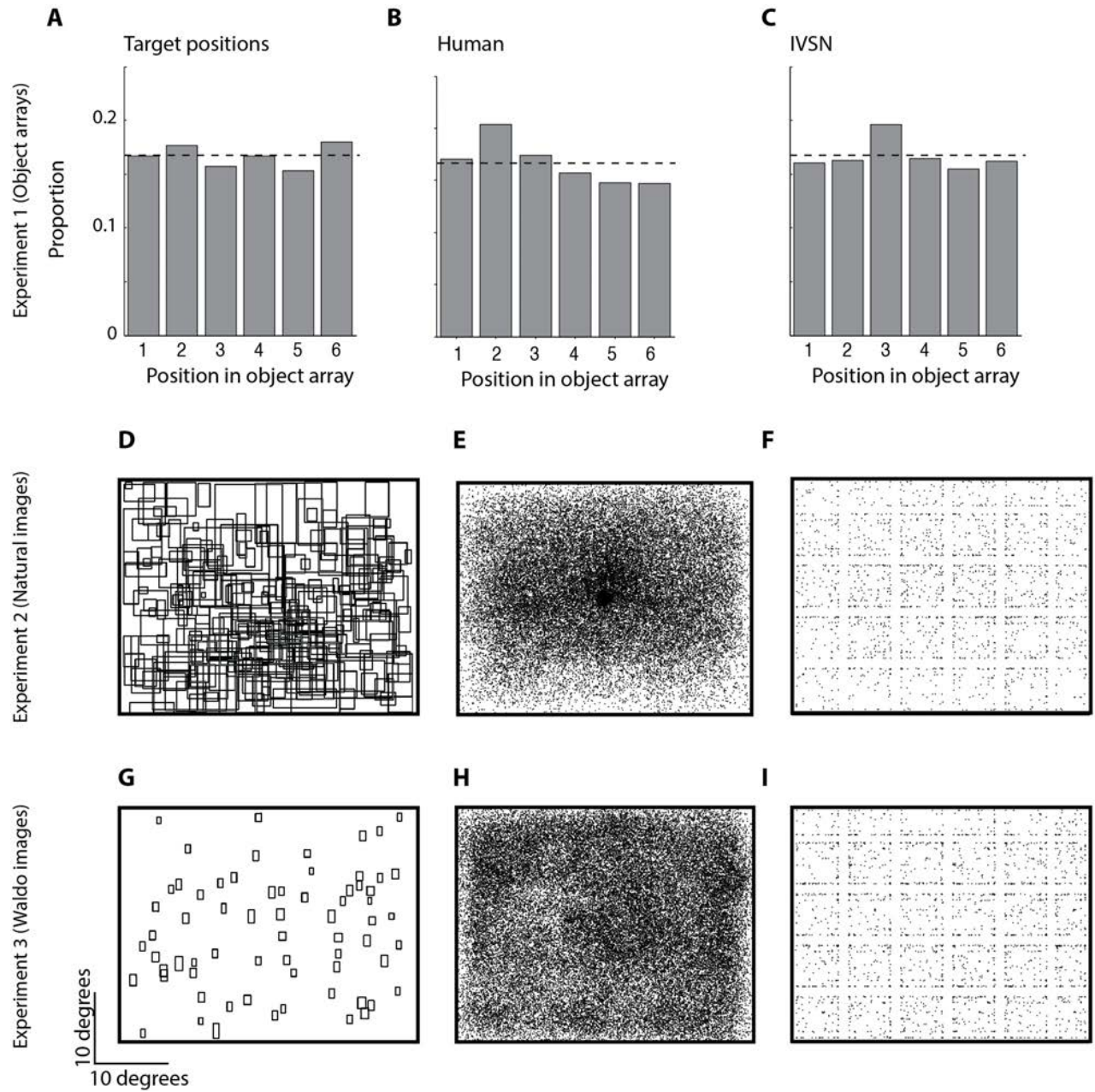

**Figure S1. Overall distribution of target locations, human fixations and model fixations.** **A, D, G.** Distribution of target locations for Experiment 1, 2 and 3, respectively. **B, E, H.** Distribution of human subject fixations for 15 subjects in each experiment (n=31,202, 99,610 and 71,346 fixations for Experiments 1, 2 and 3, respectively). In panel **H**, there is a slightly lower density of fixations in a section in the upper left quadrant; in 13/67 images, this location had text instructions and subjects were instructed to avoid this area. **C, F, I.** Distribution of IVSN fixations. The white spacing between fixations in the model is due to the way in which the large images were cropped in order to feed smaller size image segments into the model (**Methods**).

# Figure S2 A-F

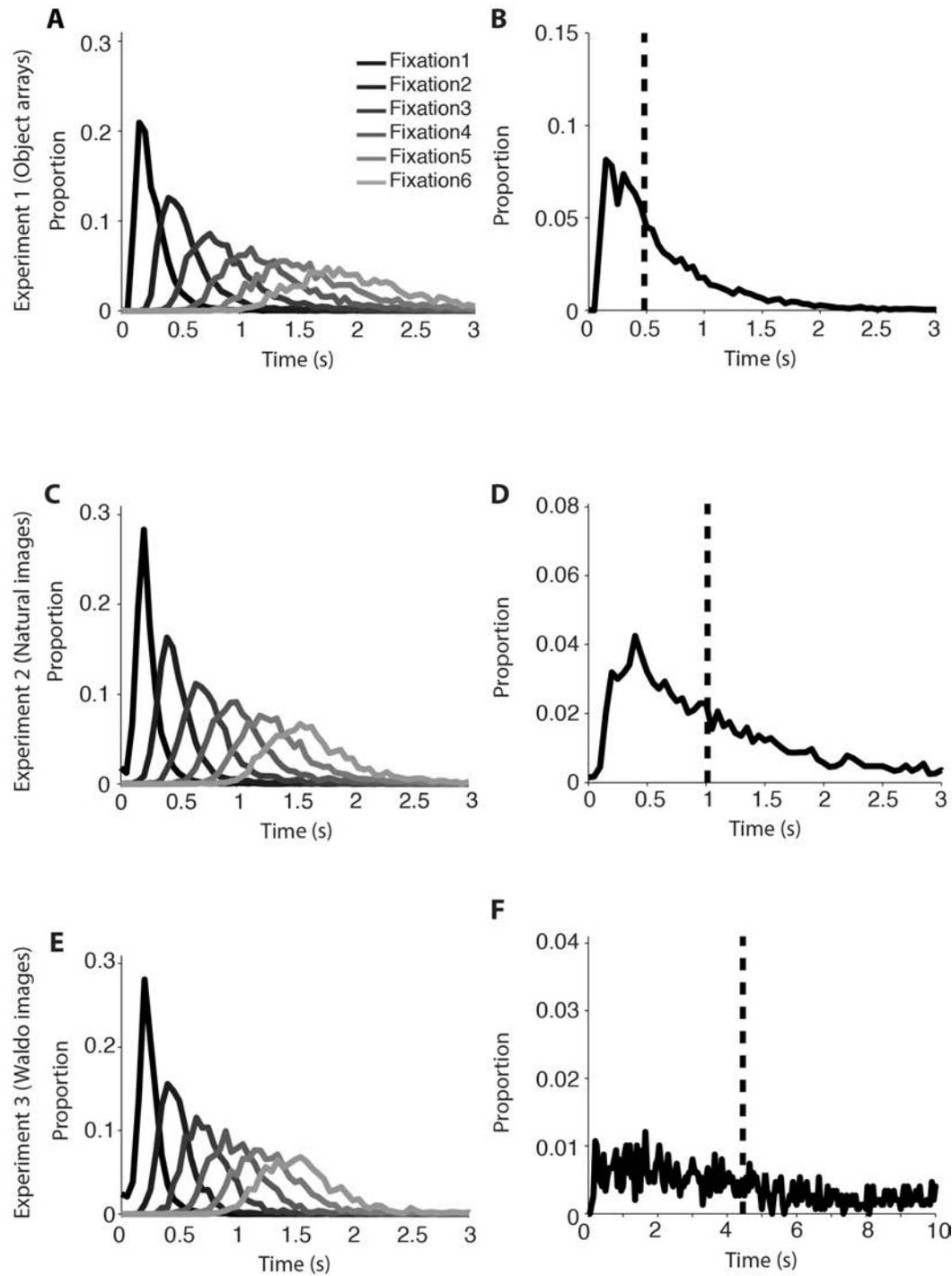

**Figure S2. Distribution of reaction times and saccade sizes.**

**A, C, E.** Distribution of reaction times for the first 6 fixations, across 15 subjects, for Experiments 1, 2 and 3, respectively. Bin size = 50 ms. In Experiments 2 and 3, there were many trials with >6 fixations (**Figure 4E**, **5E**). The distribution for the first fixation is the same as the one shown in **Figures 3D**, **4D**, and **5D** and is reproduced here for completeness. The x-axis was cut at 3 seconds.
**B, D, F.** Distribution of time required to find the target, across 15 subjects. Bin size = 50 ms. The vertical dashed line denotes the median (mean values are reported in the text). There was a significant difference in the time required to find the target among the 3 experiments (one-way ANOVA, $p<10^{-15}$, df=14217, F=3015).

# Figure S2 G-L

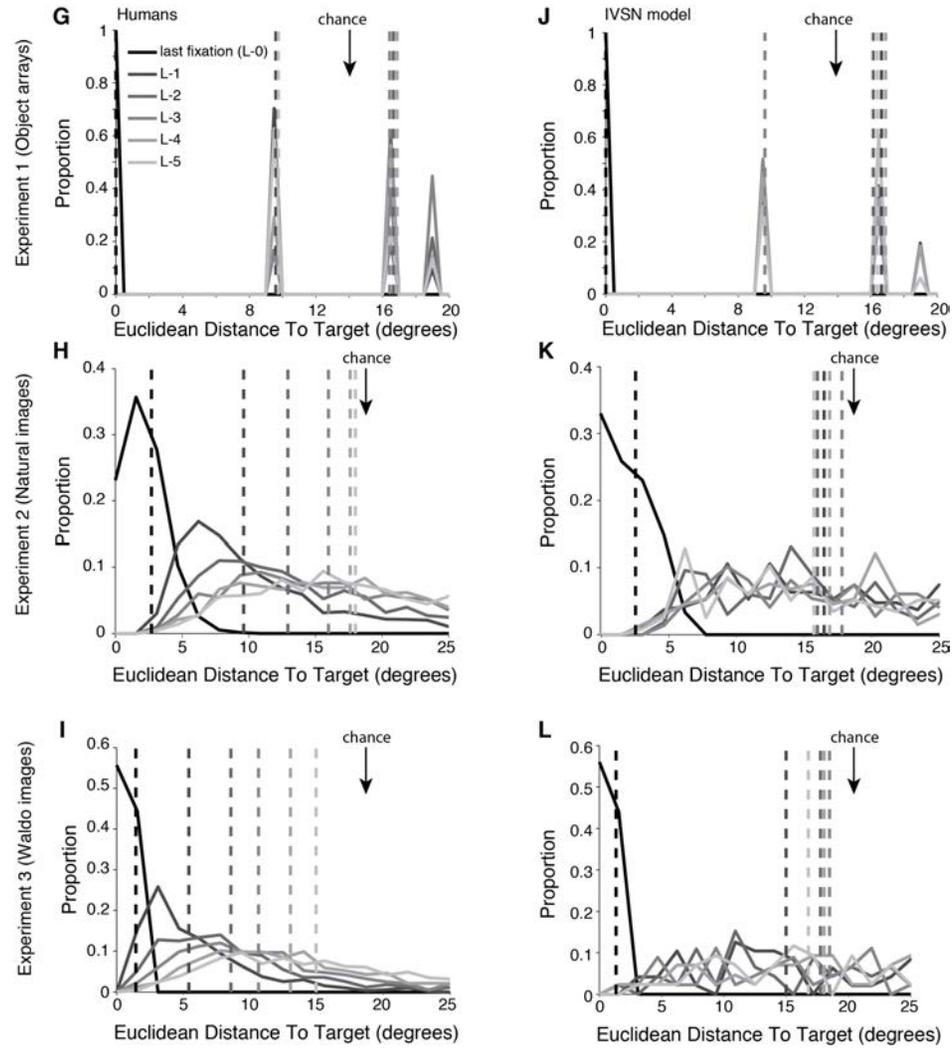

**Figure S2G-L. Distance to target for the last 6 fixations.** Distribution of the distance (in degrees of visual angle) between the fixation location and the target location for the last fixation (L-0), the fixation before last (L-1), etc. for humans (**G-I**) or the IVSN model (**J-L**). The vertical dashed lines denote the average of each distribution. On average, each subsequent fixation brought human subjects closer to the target towards the end, whereas the model was more likely to arrive at the target from a distant location. The arrows indicate the expected distance from a random location to the target. In **H, I, K, L**, given the image dimensions $L_w$=40 degrees, $L_h$=32 degrees and d=sqrt($L_w^2$+$L_h^2$), this chance distance is given by:

$$\frac{1}{15}\left(\frac{L_w^3}{L_h^2}+\frac{L_h^3}{L_w^2}+d\left(3-\frac{L_w^2}{L_h^2}-\frac{L_h^2}{L_w^2}\right)+2.5\left(\frac{L_h^2}{L_w}\ln\left(\frac{L_w+d}{L_h}\right)+\frac{L_w^2}{L_h}\ln\left(\frac{L_h+d}{L_w}\right)\right)\right)$$

Figure S3

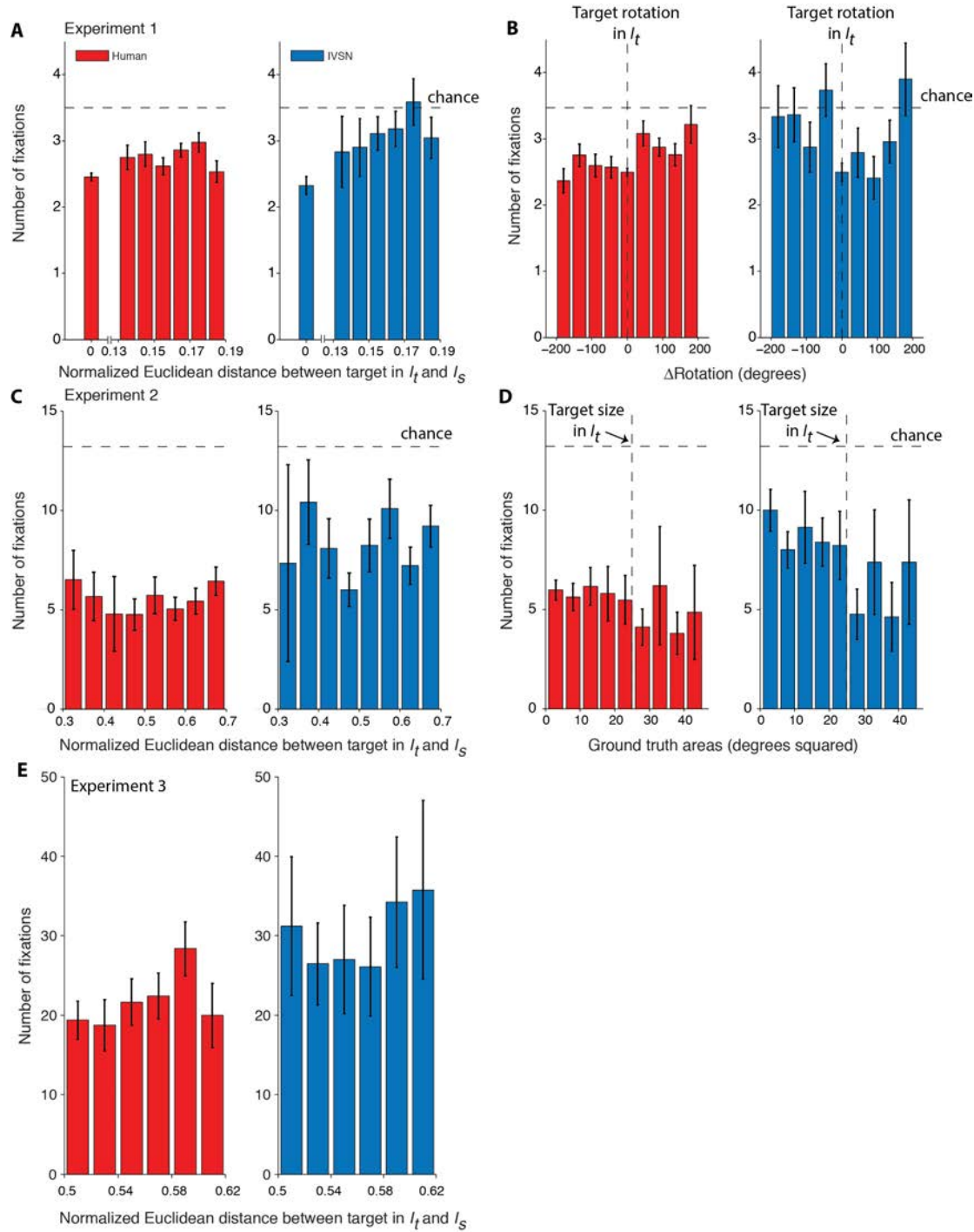

**Figure S3. Invariance in visual search.**
**A**. Number of fixations required to find the target in Experiment 1 as a function of the distance between the target as rendered in the $I_t$ and $I_s$ images. Distance = 0 corresponds to identical targets (note cut in x-axis). The horizontal dashed line indicates the null chance model.
**B.** Number of fixations required to find the target in Experiment 1 as a function of the difference between the rotation of the target object in the target image and in the search image for humans (red) and the IVSN model (blue). The vertical dashed line indicates those trials where the target was shown with the same 2D rotation angle in the $I_t$ and $I_s$ images. The horizontal dashed line indicates the null chance model.
**C**. Similar to **A** for Experiment 2. There were no distance=0 trials in this experiment.
**D**. Number of fixations required to find the target in Experiment 2 as a function of the area of the target in the $I_s$ image. The dashed line shows the size of the target object in the $I_t$ image. The horizontal dashed line indicates the null chance model.
**E**. Similar to **A** for Experiment 3. The null chance model required 58 fixations on average (beyond the y scale).

# Figure S4A

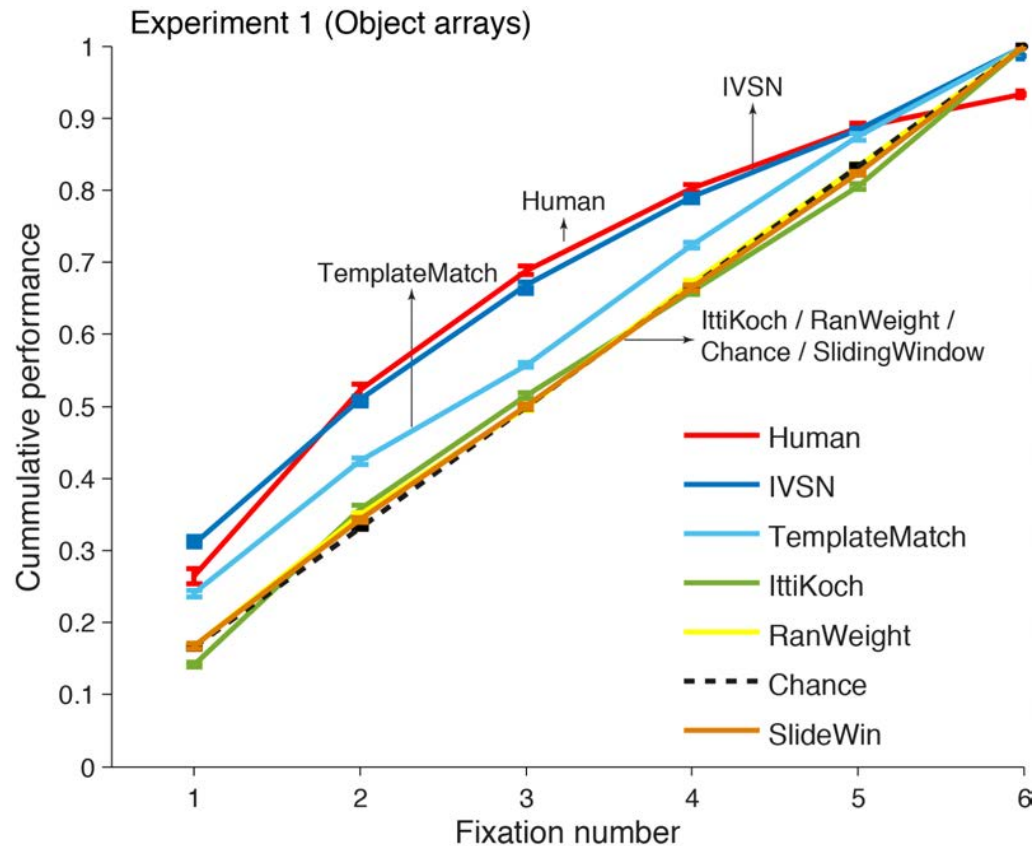

**Figure S4. Performance comparison with alternative models.** The format and conventions are the same as those in **Figures 3E**, **4E**, **5E** in the main text. Error bars denote SEM. See text and Methods for a description of each model. The curves for "Human", IVSN, and Chance are reproduced from **Figures 3E**, **4E** and **5E** for comparison purposes. **A.** Experiment 1 (Object arrays).

All models except for IVSN were statistically different from humans (two-tailed t-test):
TemplateMatching: $p<10^{-9}$,
RanWeight: $p<10^{-15}$
IttiKoch: $p<10^{-15}$
SlideWin: $p<10^{-15}$
Chance: $p<10^{-15}$
IVSN: $p=0.03$

All models were statistically different from IVSN ($p<0.01$, two-tailed t-test, df>598).
TemplateMatching: $p=0.01$
RanWeight: $p<10^{-5}$
IttiKoch: $p<10^{-6}$
SlideWin: $p<10^{-7}$
Chance: $p<10^{-12}$

# Figure S4B

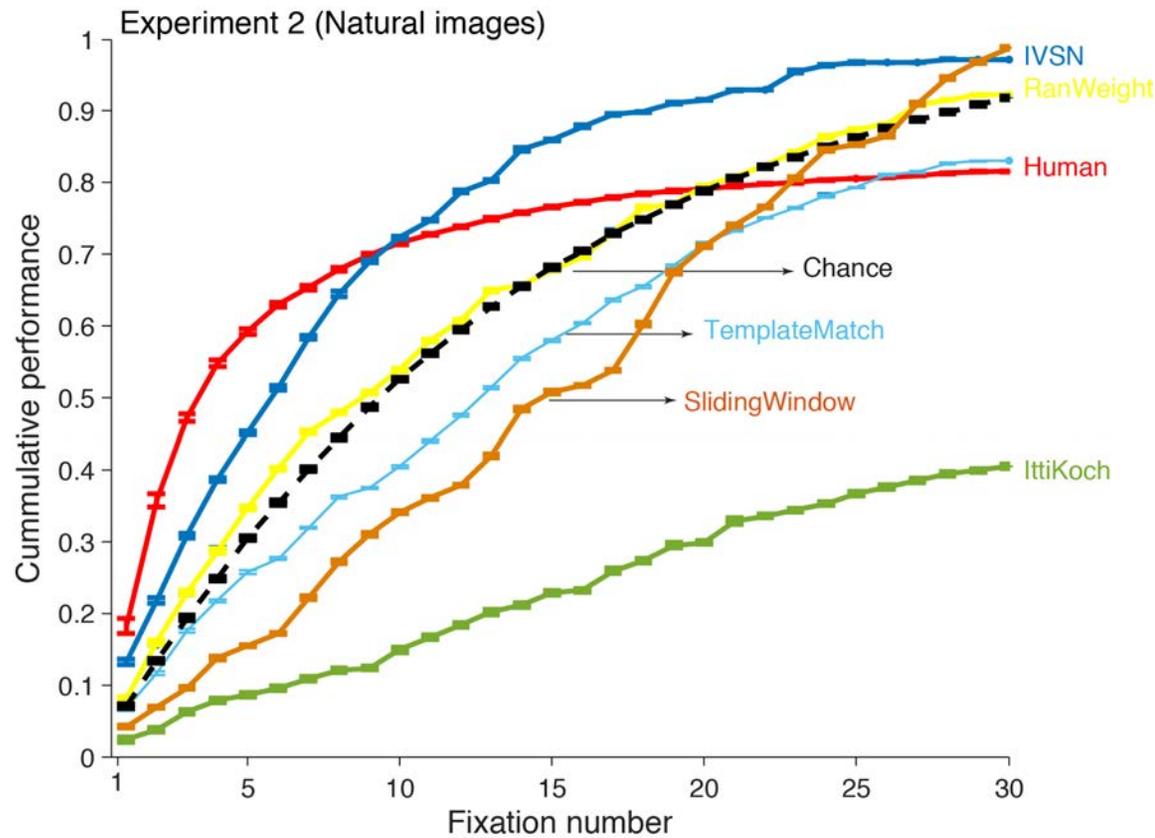

**Figure S4. Performance comparison with alternative models.**
**B**. Experiment 2 (Natural images).

All models were statistically different from humans (two-tailed t- test):
TemplateMatching: $p<10^{-15}$,
RanWeight: $p<10^{-15}$
IttiKoch: $p<10^{-15}$
SlideWin: $p<10^{-15}$
Chance: $p<10^{-15}$
IVSN: $p<10^{-5}$

All models were statistically different from IVSN (two-tailed t-test):
TemplateMatching: $p<10^{-10}$
RanWeight: $p<10^{-5}$
IttiKoch: $p<10^{-15}$
SlideWin: $p<10^{-15}$
Chance: $p<10^{-15}$

# Figure S4C

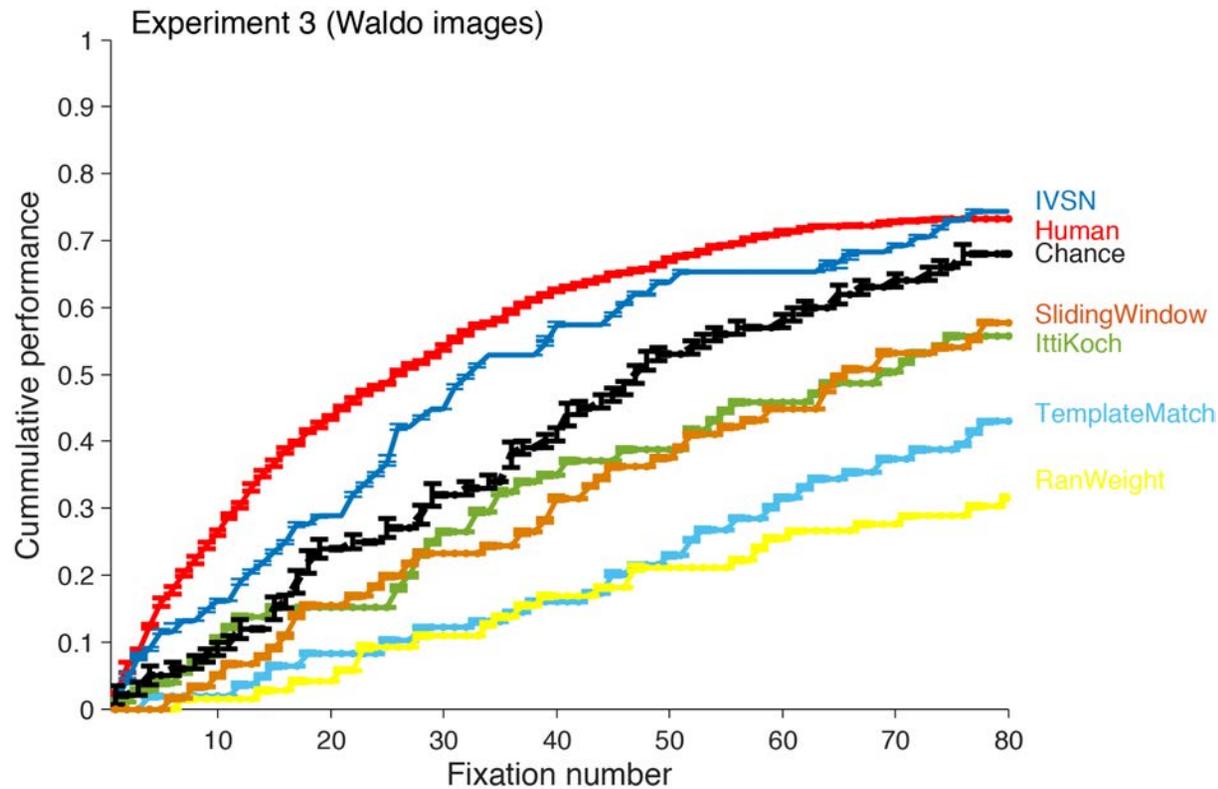

**Figure S4. Performance comparison with alternative models.**
**C**. Experiment 3 (Waldo images).

All models were statistically different from humans (two-tailed t-test):
TemplateMatching: $p<10^{-13}$,
RanWeight: $p<10^{-8}$
IttiKoch: $p<10^{-6}$
SlideWin: $p<10^{-15}$
Chance: $p<10^{-15}$
IVSN: $p=0.001$

All models were statistically different from IVSN (two-tailed t-test):
TemplateMatching: $p=0.001$
RanWeight: $p<10^{-8}$
IttiKoch: $p<0.01$
SlideWin: $p<10^{-8}$
Chance: $p<10^{-15}$

# Figure S5

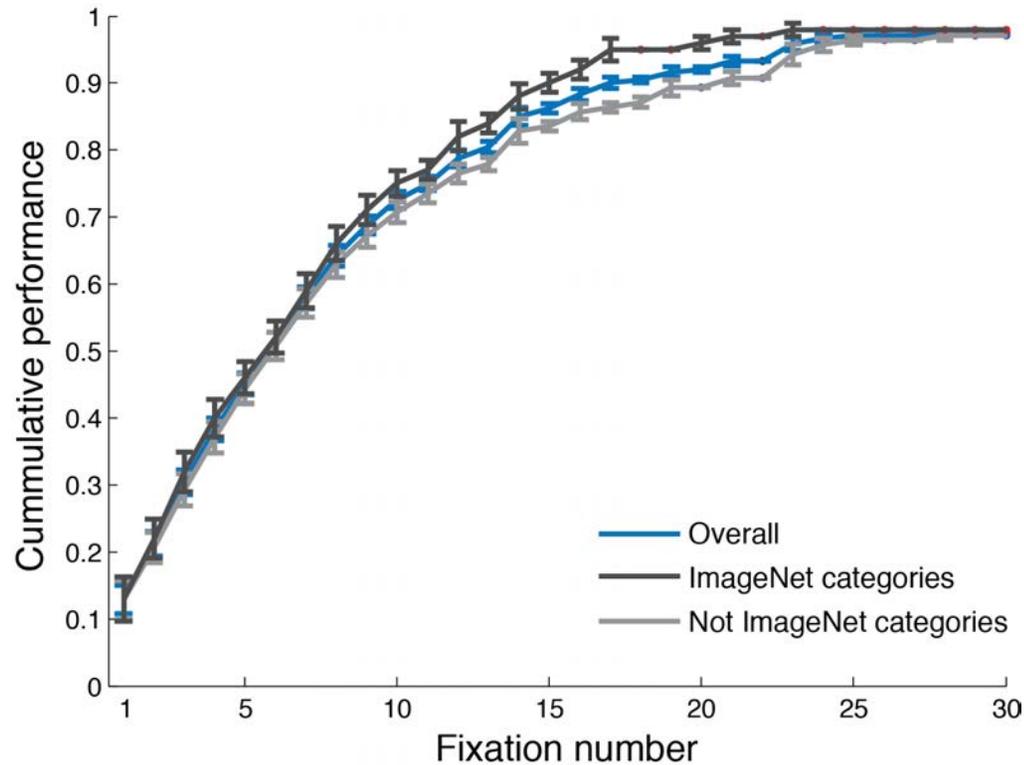

**Figure S5. Performance for ImageNet categories and non-ImageNet categories in Experiment 2.** Following the format in **Figure 4E**, cumulative performance as a function of fixation number for all images (blue, same copied from **Figure 4E**), 100 images with target object categories that were within ImageNet (dark gray) and 140 that did not (light gray). Although performance was slightly higher for target objects in ImageNet categories, there was no significant difference between the number of fixations required to find the target for ImageNet or non-ImageNet images (p=0.25, two-tailed t-test, t=1.2, df=238).

# Figure S6

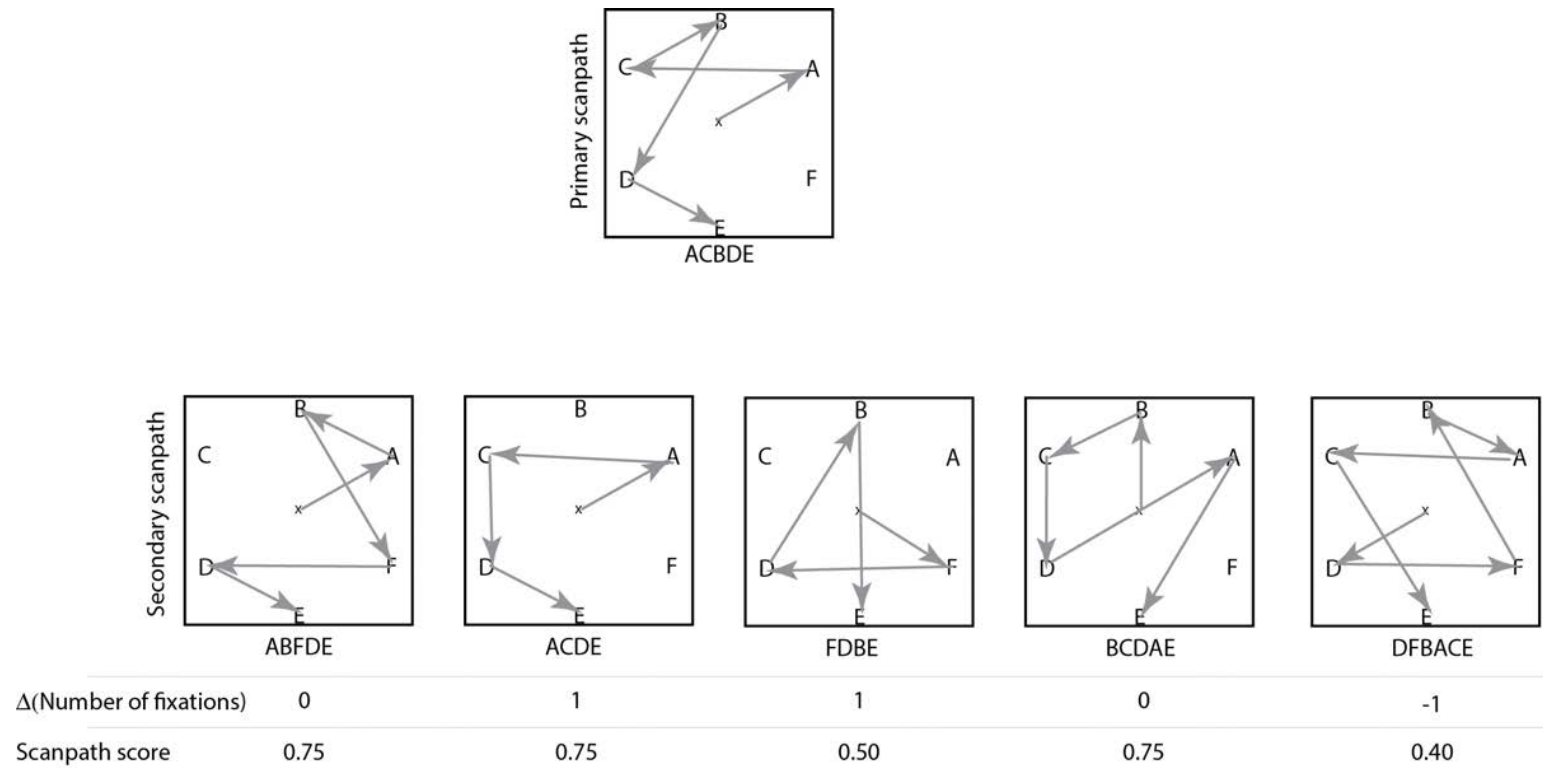

**Figure S6. Illustration of image-by-image consistency metrics in fixation patterns.** This schematic shows a comparison between a primary scan path (top, sequence = ACBDE) and alternative scan paths (middle) in a search image consisting of 6 objects where the target is at location E. The numbers below each subplot show the difference in the number of fixations and the scan path similarity score for each comparison with the primary scan path (**Methods**).

# Figure S7A

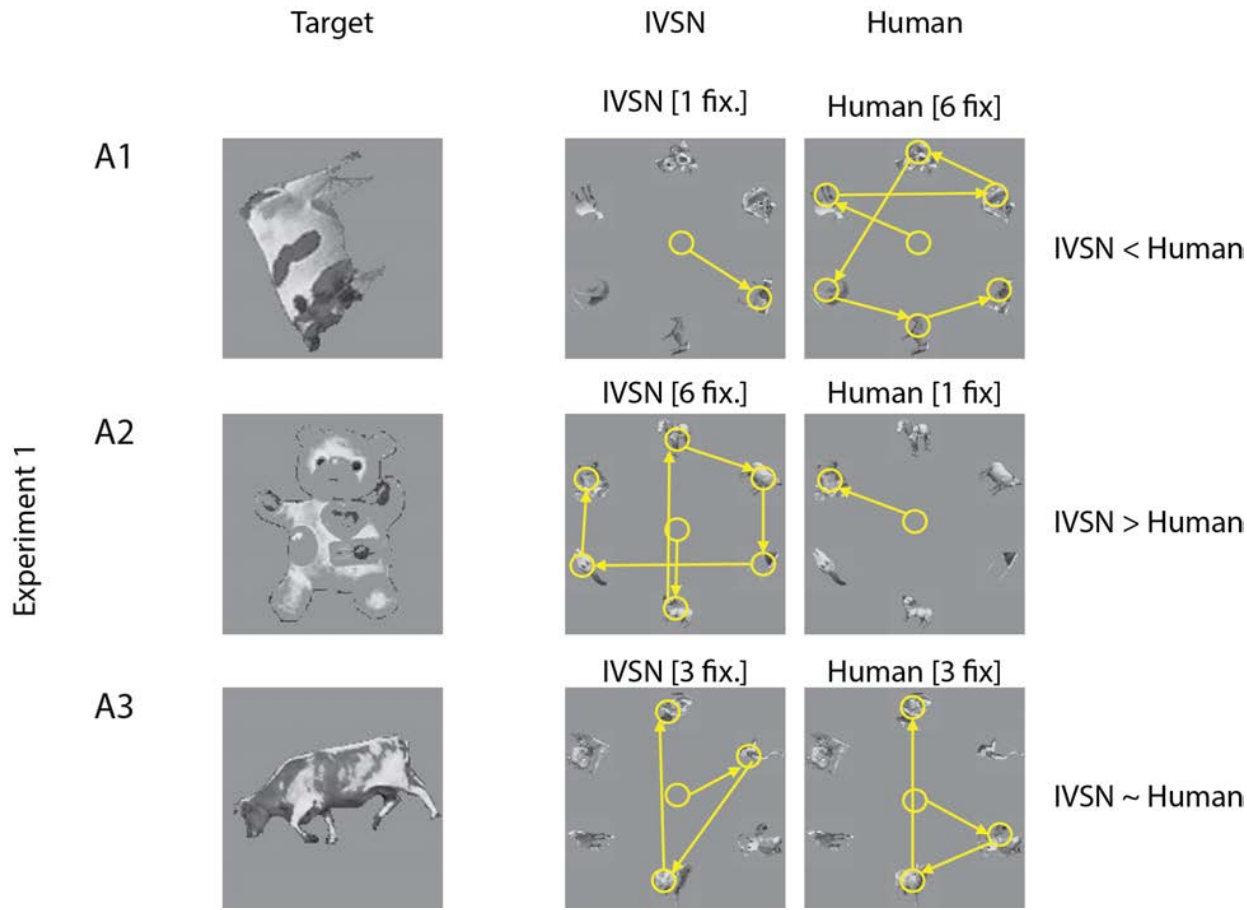

**Figure S7. Image-by-image comparison of number of fixations required to find the target. A-C.** Example trials where the IVSN model found the target faster than humans (**A1, B1, C1**), when humans found the target faster than the IVSN model (**A2, B2, C2**), and trials where humans and the IVSN model were comparable (**A3, B3, C3**) for Experiment 1 (**A**), Experiment 2 (**B**), and Experiment 3 (**C**). The left column shows the target image, columns 2 and 3 show the sequence of fixations for the IVSN model (column 2) and one of the subjects (column 3). The number of fixations required to find the target is shown above each search Image.

# Figure S7B

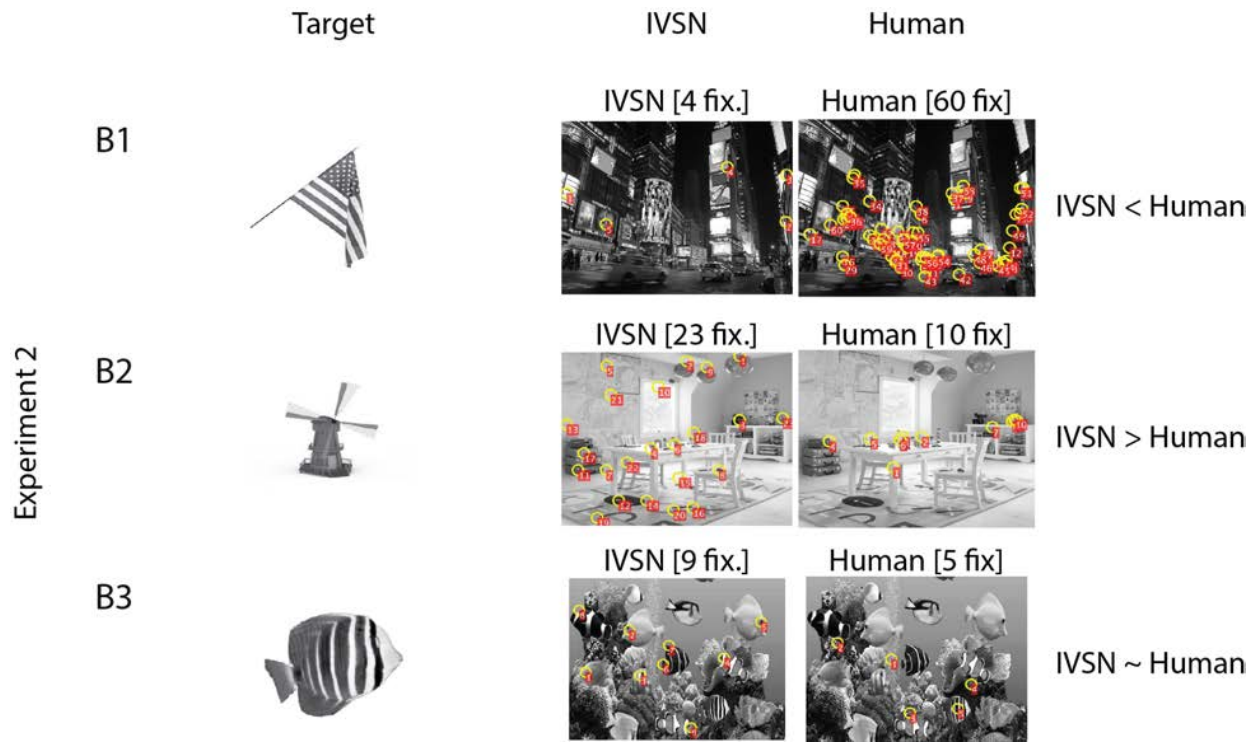

**Figure S7. Image-by-image comparison of number of fixations required to find the target.** See previous panel for legend.

# Figure S7C

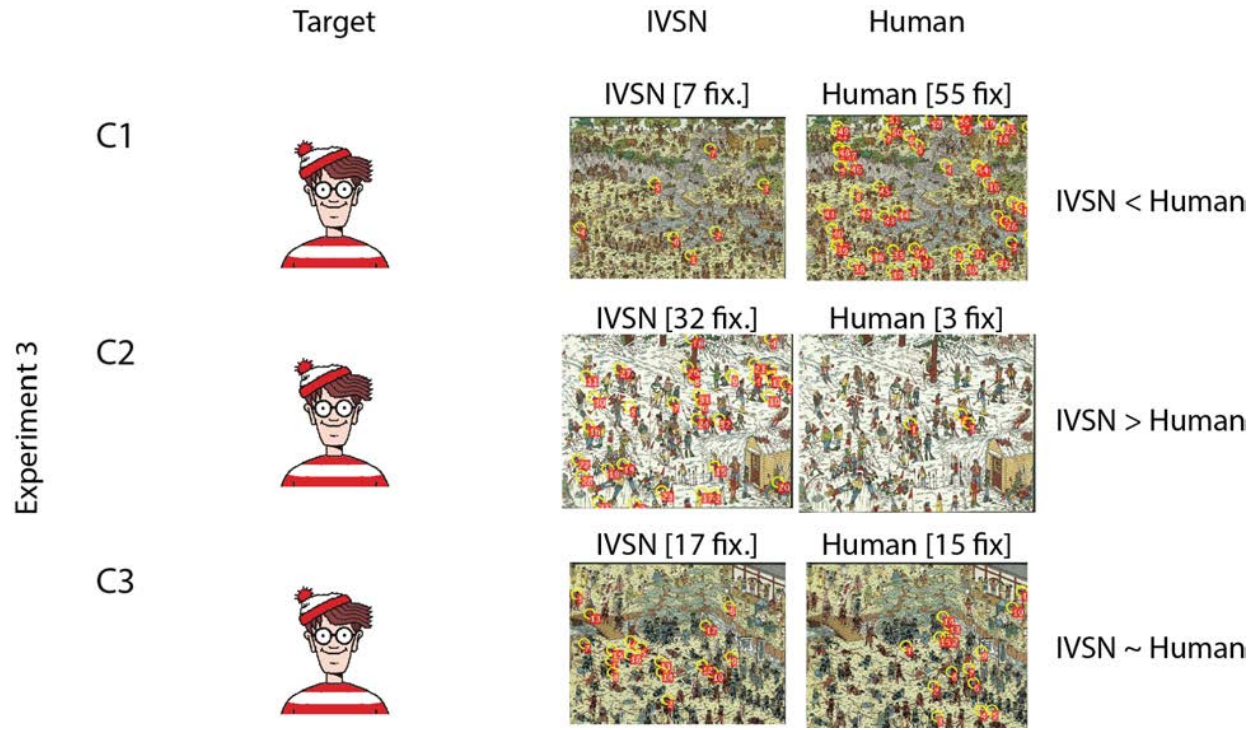

**Figure S7. Image-by-image comparison of number of fixations required to find the target.** See previous panel for legend.

# Figure S7DEF

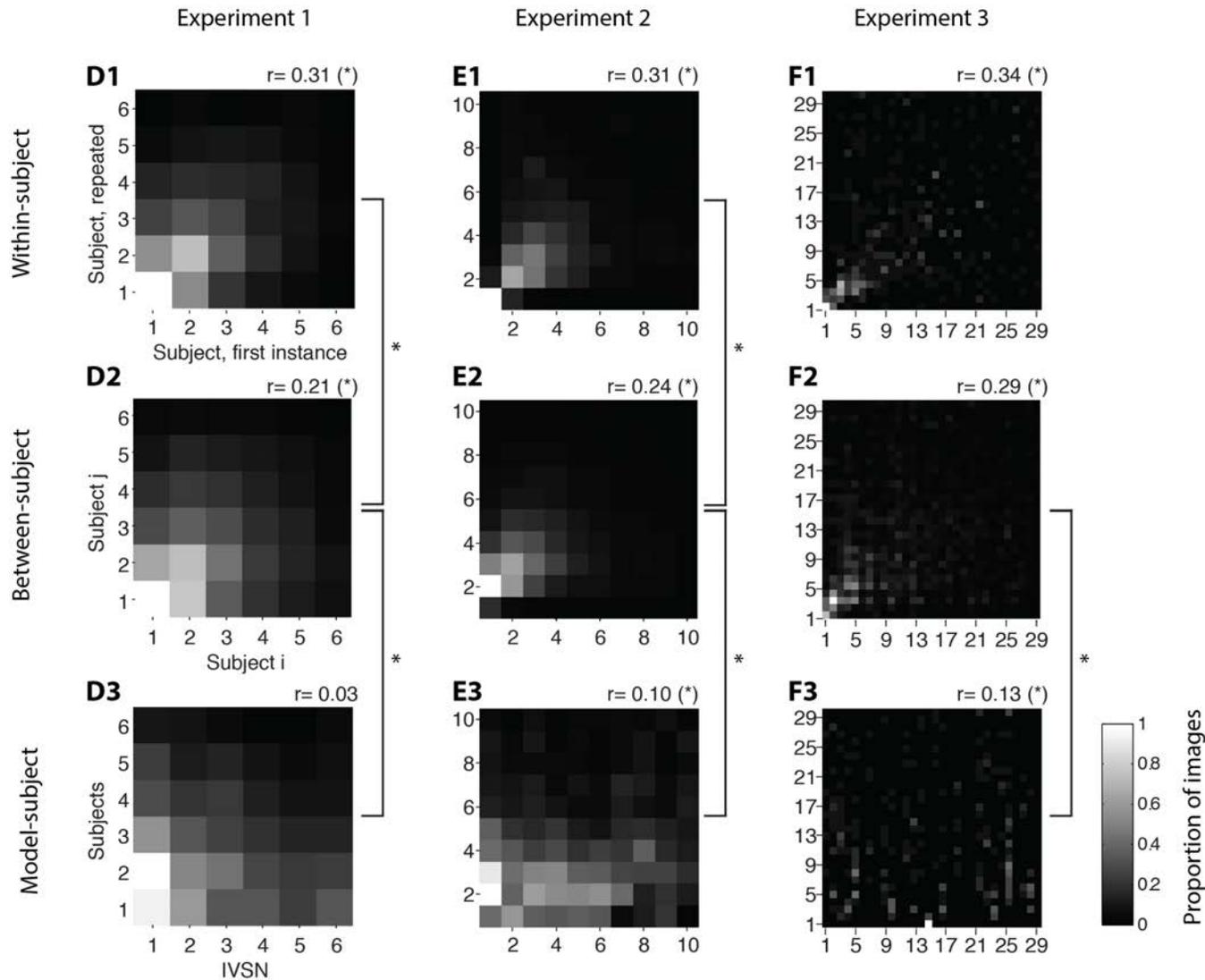

**Figure S7. Image-by-image comparison of number of fixations required to find the target. D-F.** Comparison in the number of fixations required to find the target averaged across subjects for each experiment (columns), within subjects (**D1**, **E1**, **F1**), between subjects (**D2**, **E2**, **F2**) and between subjects and IVSN model (**D3**, **E3**, **F3**). When comparing S1 and S2 (e.g., two subjects), entry (*i,j*) indicates the proportion of images where S1 required *i* fixations and S2 required *j* fixations (see scale bar on bottom right). Presence of entries exclusively along the diagonal would indicate that the behavior of S1 and S2 is identical on an image-by-image basis. Results were averaged across subjects (see **Figures S7G-I** for distribution for individual subjects). Note that the size of the matrices are different for each experiment, reflecting the increasing difficulty from Experiment 1 to 3. The r values show the average of the correlation coefficients computed in **Figures S7G-I** in the subject-by-subject comparisons. An * next to the *r* value indicates that the distribution of *r* values was different from zero (two-tailed t-test, $p<0.01$). An * comparing two matrices indicates that the distributions of *r* values were statistically different (two-tailed t-test, $p<0.01$).

# Figure S7GHI

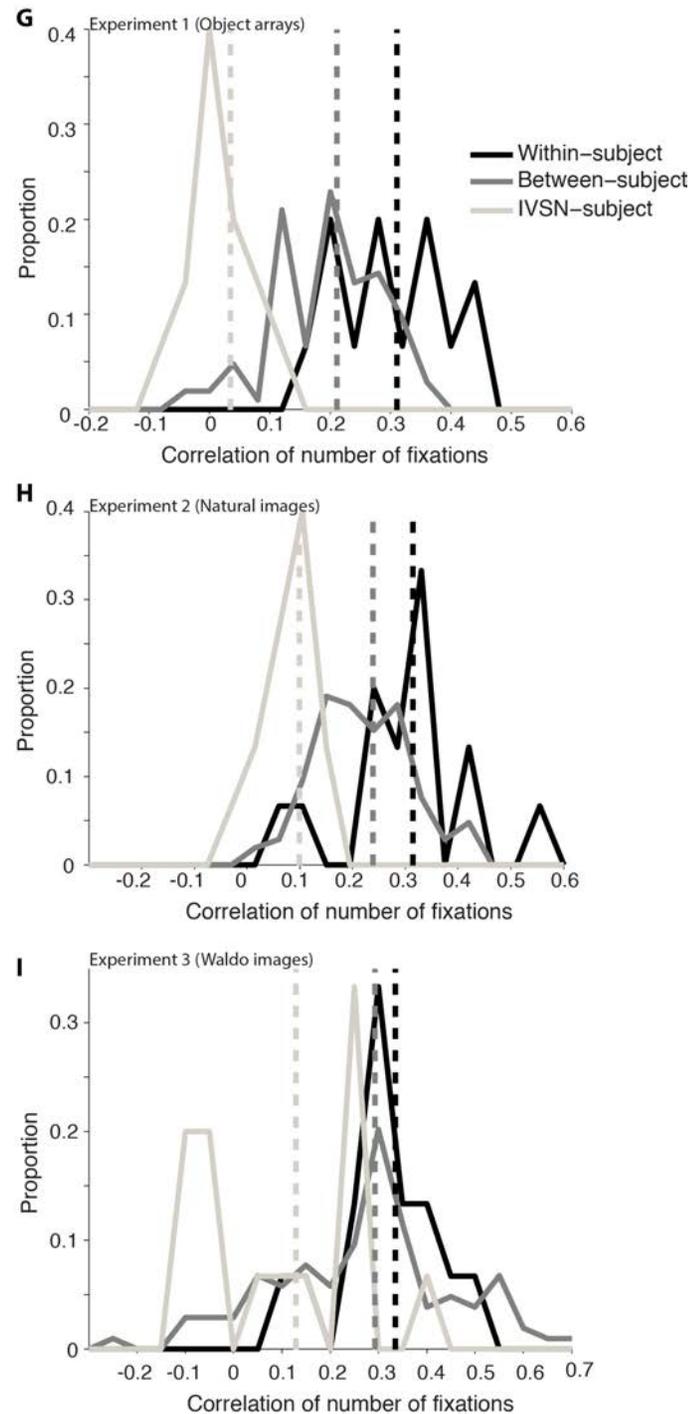

**Figure S7. Image-by-image comparison of number of fixations required to find the target.**
Using the same comparison of the number of fixations described for **Figure S7DEF**, this figure shows the distribution of the correlation coefficients on a subject-by-subject basis for Experiment 1 (**G**), Experiment 2 (**H**) and Experiment 3 (**I**). The colors denote the within-subject comparisons (black), between subject comparisons (dark gray), and IVSN-subject comparisons (light gray).

# Figure S8

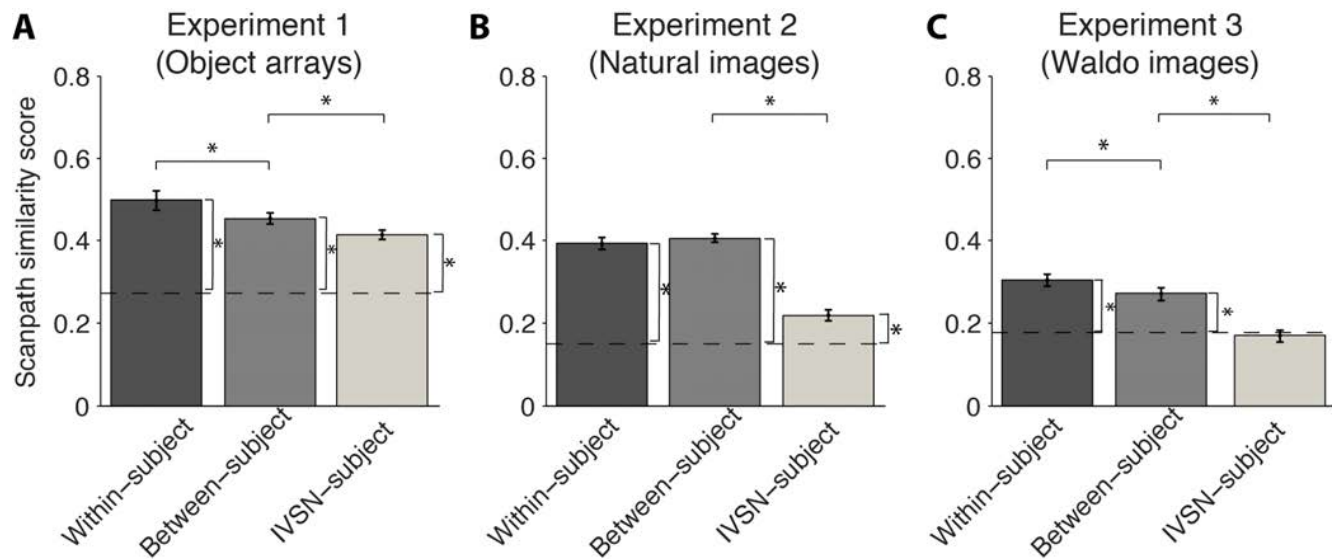

**Figure S8: Image-by-image consistency in the spatiotemporal pattern of fixation sequences using entire fixation sequences.** Scanpath similarity scores (see text and **Methods** for definition) comparing the fixation sequences within subjects (dark gray), between-subjects (medium gray) and between the IVSN model and subjects (light gray) for Experiment 1 (**A**), Experiment 2 (**B**), and Experiment 3 (**C**). The larger the scanpath similarity score, the more similar the fixation sequences are. The dashed line indicates chance performance, obtained by randomly permuting the images. Results shown here are averaged over subjects and subject pairs. The "*" denote statistical significance (p<0.01, two-tailed t-test), comparing each result against chance levels (vertical comparisons) and comparing within-subject versus between-subject scores and between-subject versus IVSN-subject scores (horizontal comparisons).

# Figure S9

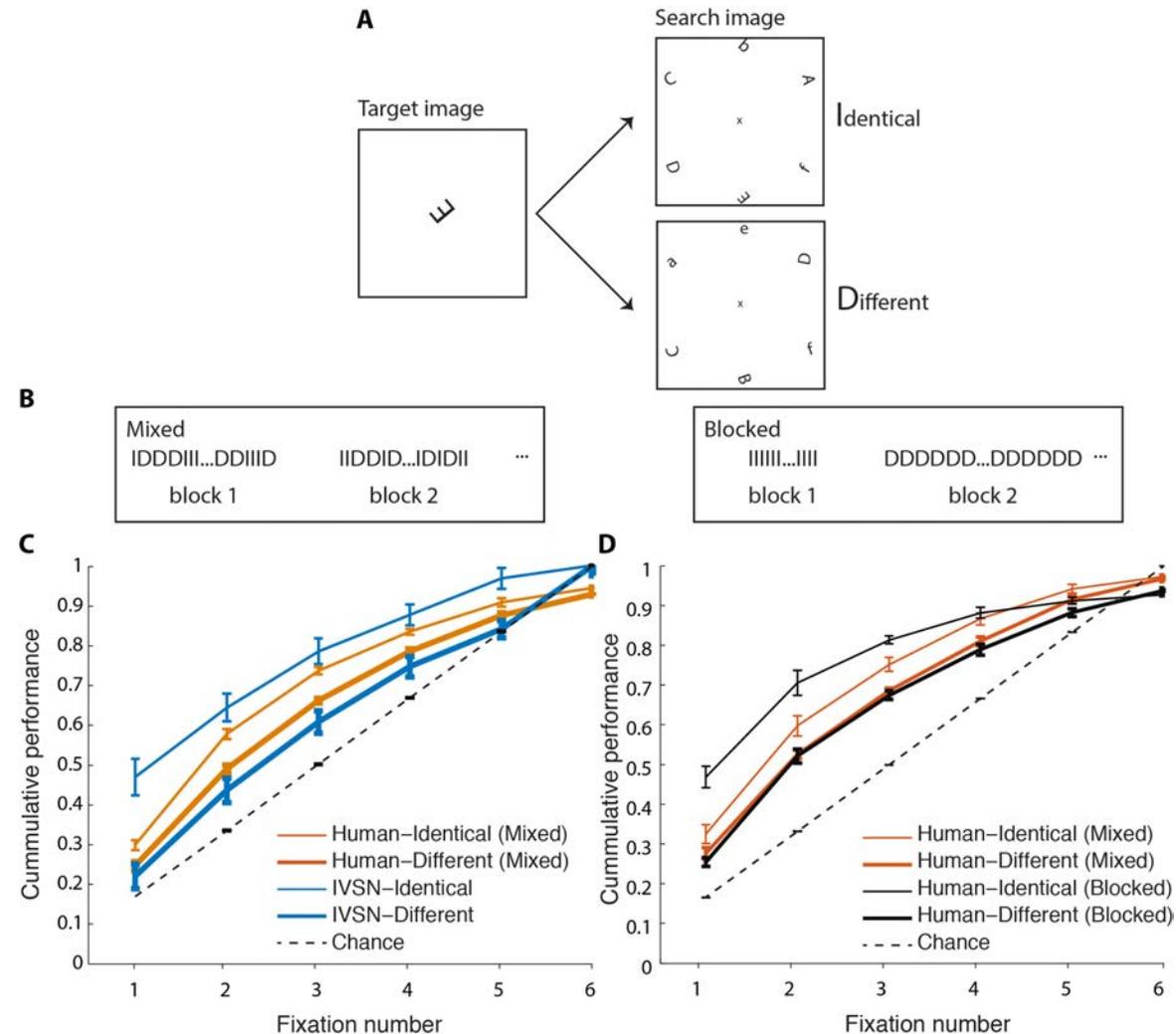

**Figure S9. Blocked identical trials yielded improved performance (Experiment 1). A.** The target as rendered in the target image could be identical (I) to the one in the search image or different (D). **B.** In the mixed condition, all trials were randomized (left). In the blocked condition, all the trials within a block consisted of the target identical condition or the target different condition (right). **C.** In the target identical condition (thin lines), there was an improvement in performance both for humans (red, $p<10^{-7}$, two-tailed t-test, t=5.6, df=4173) and the IVSN model (blue, $p<10^{-5}$, two-tailed t-test t=4.6, df=298) compared to the target different condition (thick lines). **D.** During the experiments reported in the main text, trial order was randomized (Mixed, red). We conducted a separate experiment where trials were blocked such that all Identical trials were together and all Different trials were together (Blocked, black). Within the blocked trials, performance was higher in the Identical trials ($p<10^{-21}$, two-tailed t-test, t=9.8, df=1398). In addition, performance in Identical blocked trials was better than performance in Identical mixed trials ($p<10^{-14}$, two-tailed t-test, t=7.9, df=2236). In contrast, there was no significant difference between the Different blocked trials and the Different mixed trials (p=0.49, two-tailed t-test, t=0.69, df=3335).

# Figure S10

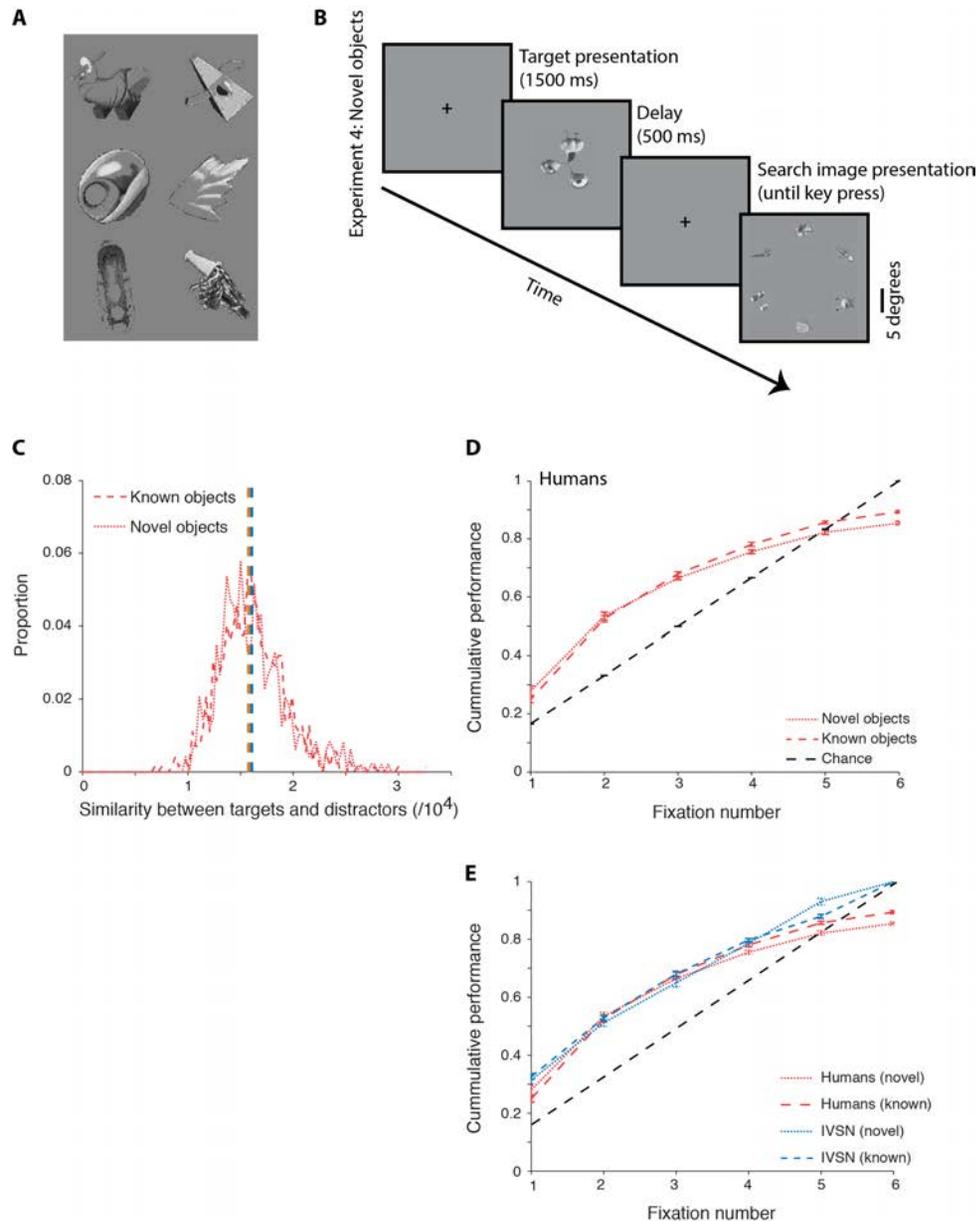

**Figure S10. Humans can find novel objects.**
**A.** Six example novel objects out of the 1860 novel objects from 98 categories.
**B**. Schematic of Experiment 4. The novel objects experiment followed the structure of Experiment 1.
**C**. Difficulty match between known objects (those from Experiment 1) and novel objects. The distribution of similarity scores between targets and distractors for all trials was similar for known objects and novel objects (**Methods**, p>0.6, t=-0.5, df=1204).
**D**. Cumulative performance following the same format as **Fig. 3E** for known objects (dashed line) and novel objects (dotted line). Performance for both novel and known objects was above chance (p<$10^{-15}$ and p<$10^{-15}$, respectively). There was a small, but significant, difference in performance between novel and known objects (average number of fixations: 2.42+/-1.43 and 2.54+/-1.42, respectively, p=0.004, t=2.9, df=5278, two-tailed t-test).

**E**. IVSN model performance for known objects (dashed blue) and novel objects (dotted blue). Human performance is copied from part **D** for comparison. IVSN performance for both novel and known objects was above chance (p<$10^{-17}$ and p<$10^{-12}$, respectively).

The novel objects were collected from the following sources:
1. Horst, J. S., & Hout, M. C. The Novel Object and Unusual Name (NOUN) Database: A collection of novel images for use in experimental research. Behavior Research Methods, 2016. Retrieved from: http://michaelhout.com/?page_id=759.
2. Michael Tarr's web site for Freebles, Greebles, Yadgits, YUFOs: http://wiki.cnbc.cmu.edu/Novel_Objects
3. Alien 3D models:
https://www.turbosquid.com/Search/Index.cfm?keyword=alien&max_price=0&min_price=0

# Figure S11 A-C

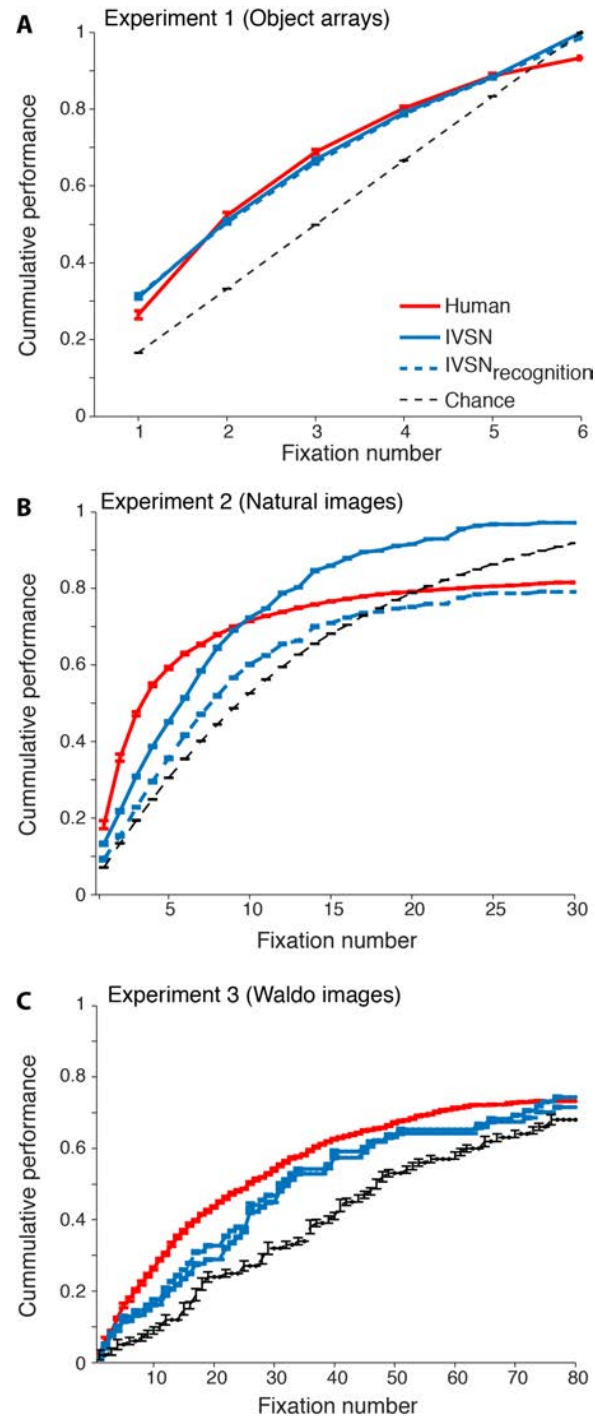

**Figure S11. Object recognition, memory and saccade sizes. A-C.** The results presented in the main text use an "oracle" to determine whether the target is present at a given location or not (**Methods**). Here we introduce a recognition mechanism into the model (IVSN$_{recognition}$) to determine whether the target is present at a given location or whether the model should continue search. These figures match **Figures 3E**, **4E** and **5E** (the red, blue and black dashed lines are copied from those figures for comparison purposes) and introduces the dashed blue line model (IVSN$_{recognition}$).

The performance of the IVSN$_{recognition}$ model was different from humans in Experiment 2 (two-tailed t-test):
Experiment 1: p=0.04
Experiment 2: p<$10^{-5}$
Experiment 3: p=0.02

The performance of the IVSN$_{recognition}$ model was significantly better than chance (two-tailed t-test):
Experiment 1: p<$10^{-15}$
Experiment 2: p<$10^{-13}$
Experiment 3: p<$10^{-15}$

# Figure S11 D-F

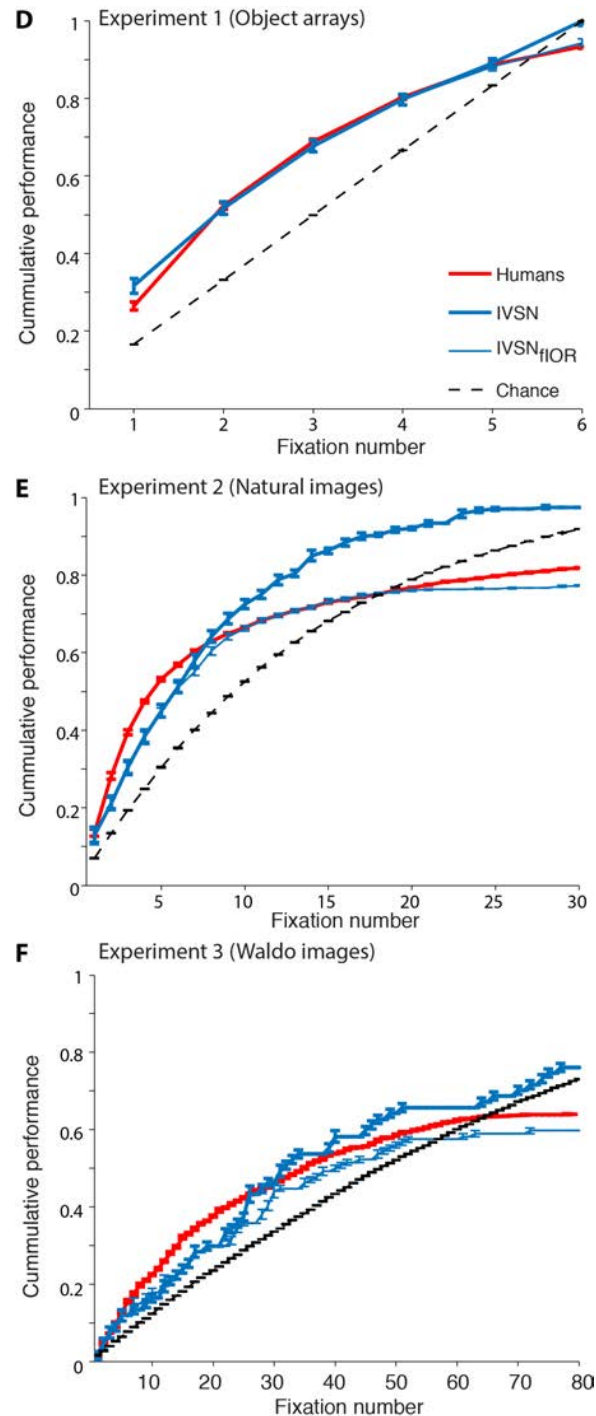

**Figure S11. Object recognition, memory and saccade sizes. D-F.** The model presented in the main text assumes infinite inhibition of return. Here we introduce finite inhibition of return into the model (**Methods**, IVSN$_{fIOR}$. These figures match **Figures 3E**, **4E** and **5E** (the red, blue and black dashed lines are copied from those figures for comparison purposes) and introduces the thin blue line model (IVSN$_{fIOR}$).

The performance of the IVSN$_{fIOR}$ model was not different from humans (two-tailed t-test):
Experiment 1: p=0.87
Experiment 2: p=0.027
Experiment 3: p=0.29

The performance of the IVSN$_{fior}$ model was significantly better than chance (two-tailed t-test):
Experiment 1: $p<10^{-15}$
Experiment 2: $p<10^{-15}$
Experiment 3: $p<10^{-15}$

# Figure S11G-L

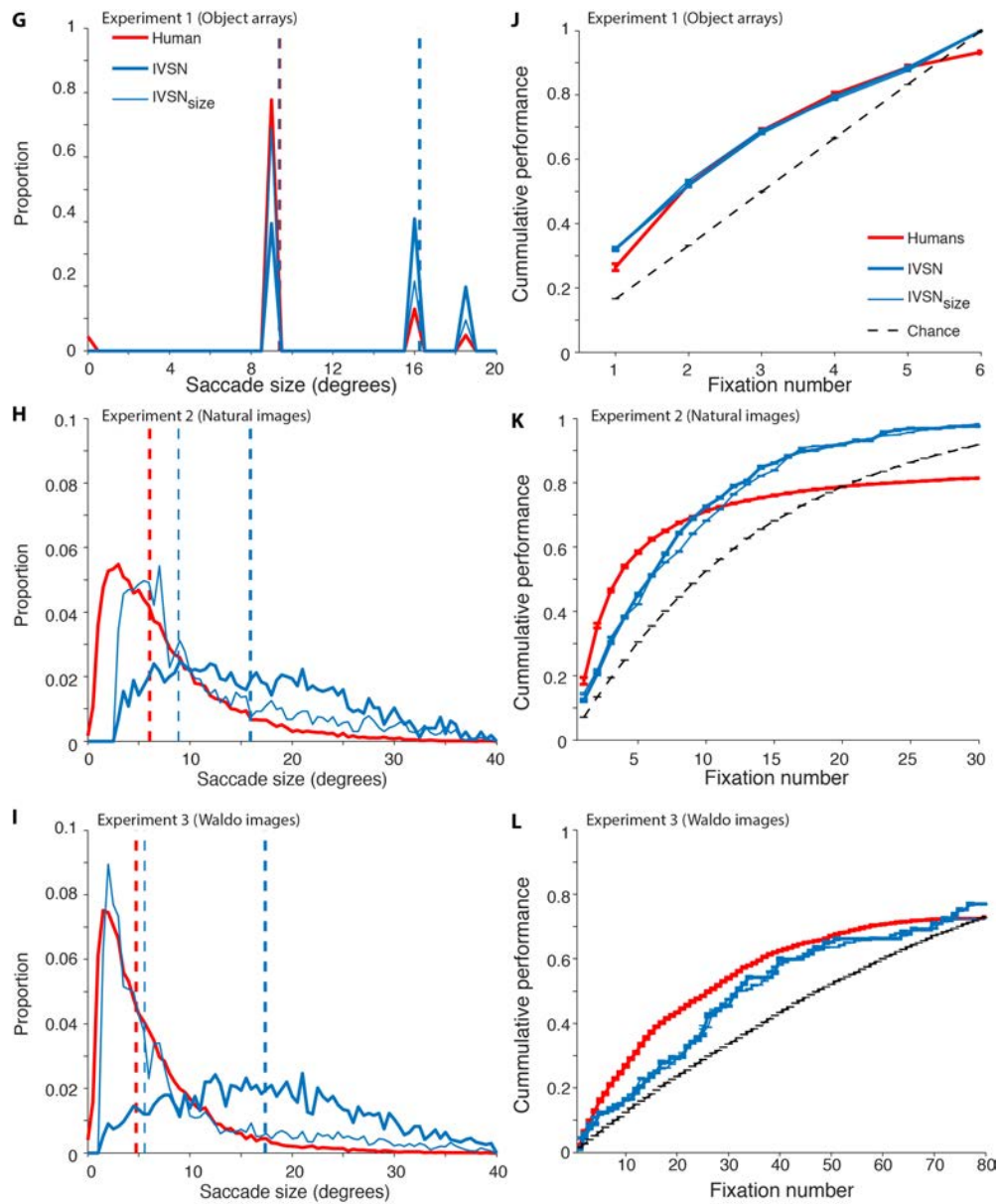

**Figure S11. Object recognition, memory and saccade sizes. G-I.** Distribution of saccade sizes for Experiments 1, 2 and 3, respectively, for humans (red), the IVSN model (thick blue), and the IVSN model constrained by saccade distance (IVSN$_{size}$, thin blue). The vertical dashed lines show the median values. In all experiments, there was a significant difference between humans and the IVSN model ($p<10^{-15}$, two-tailed t-test, $t>23$).

**J-L. Performance of the IVSN$_{size}$ model.** The format is the same as that in **Figures 3E**, **4E** and **5E** and the red, thick blue, and black dashed lines are copied from those figures for comparison purposes.

The performance of the IVSNsize model was significantly different from humans (two-tailed t-test):
Experiment 1: $p=0.004$
Expeirment 2: $p<10^{-12}$
Experiment 3: $p=0.002$

The performance of the IVSNsize model was significantly different from chance (two-tailed t-test):
Experiment 1: $p<10^{-11}$
Expeirment 2: $p<10^{-14}$
Experiment 3: $p<10^{-15}$

# Figure S12

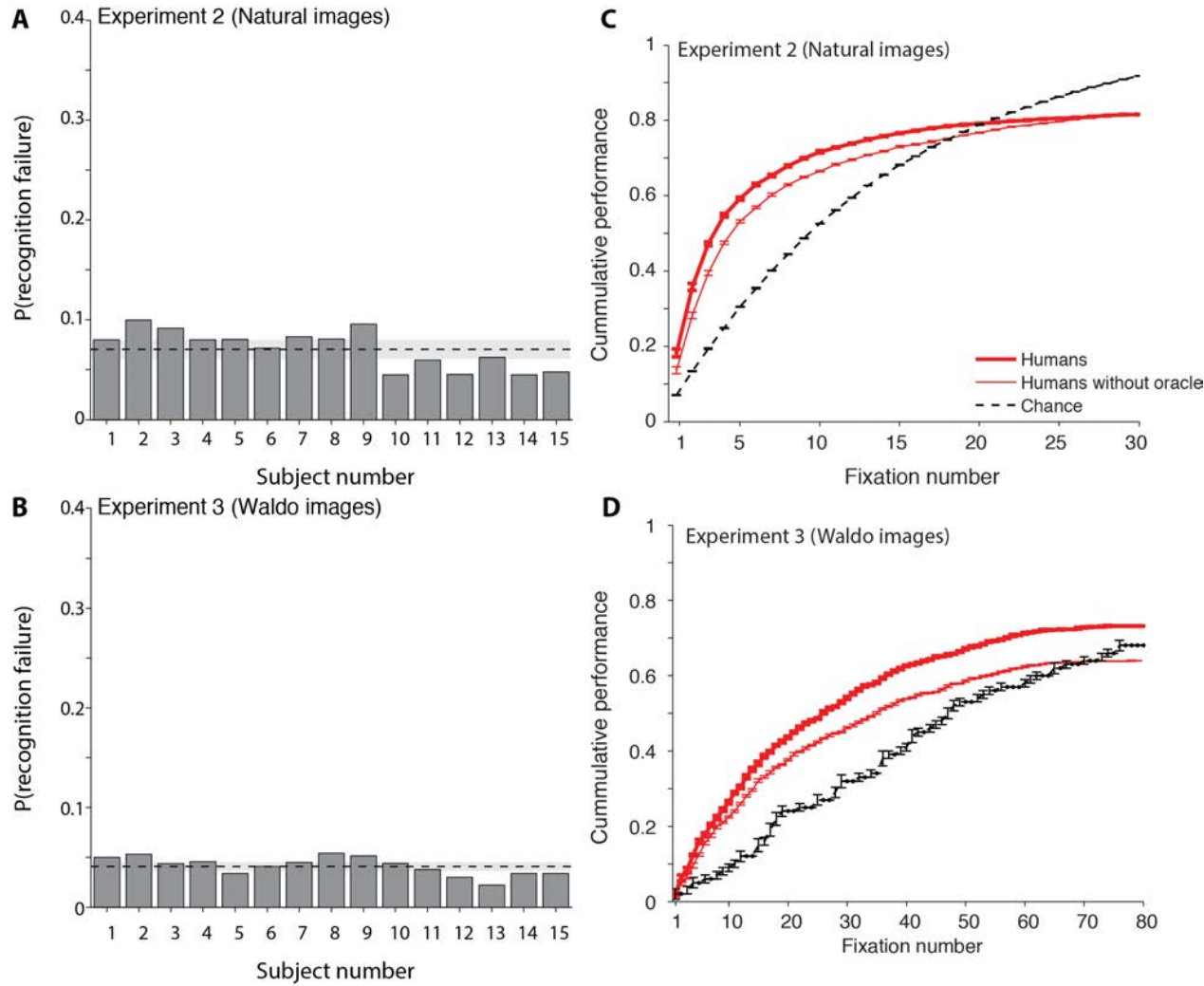

**Figure S12. Humans may fixate on the target but fail to recognize it.**
**A-B.** Probability of fixating on the target and failing to recognize it (not clicking on the target location with the mouse and continuing visual search) for each of the 15 subjects. The dashed line shows the average across subjects; the shaded area is one SD.
**C-D.** To directly compare the model and humans, the results presented throughout the text use an oracle to determine whether the target was found or not (except for IVSN$_{recognition}$ in **Figure S11A-C**). If a fixation landed on the target, the target was deemed to be found. In Experiments 2 and 3 -- but not in Experiment 1 -- subjects were asked to indicate the target location with the mouse. Here we compare performance using the oracle version (Humans, thick red line, copied from **Figures 4E** and **5E**) versus performance determined by the time when subjects click the mouse (Humans without oracle, thin red line) for Experiment 2 (**A**) and Experiment 3 (**B**). Human performance without the oracle was also above chance in both cases (p<10$^{-15}$ and p<10$^{-15}$ in **C**, **D**). Human performance with the oracle was different from that without the oracle in Experiment 2 (p<10$^{-15}$, t=11, df=6156), but not in Experiment 3 (p=0.62, t=0.49, df=1375).

# Figure S13

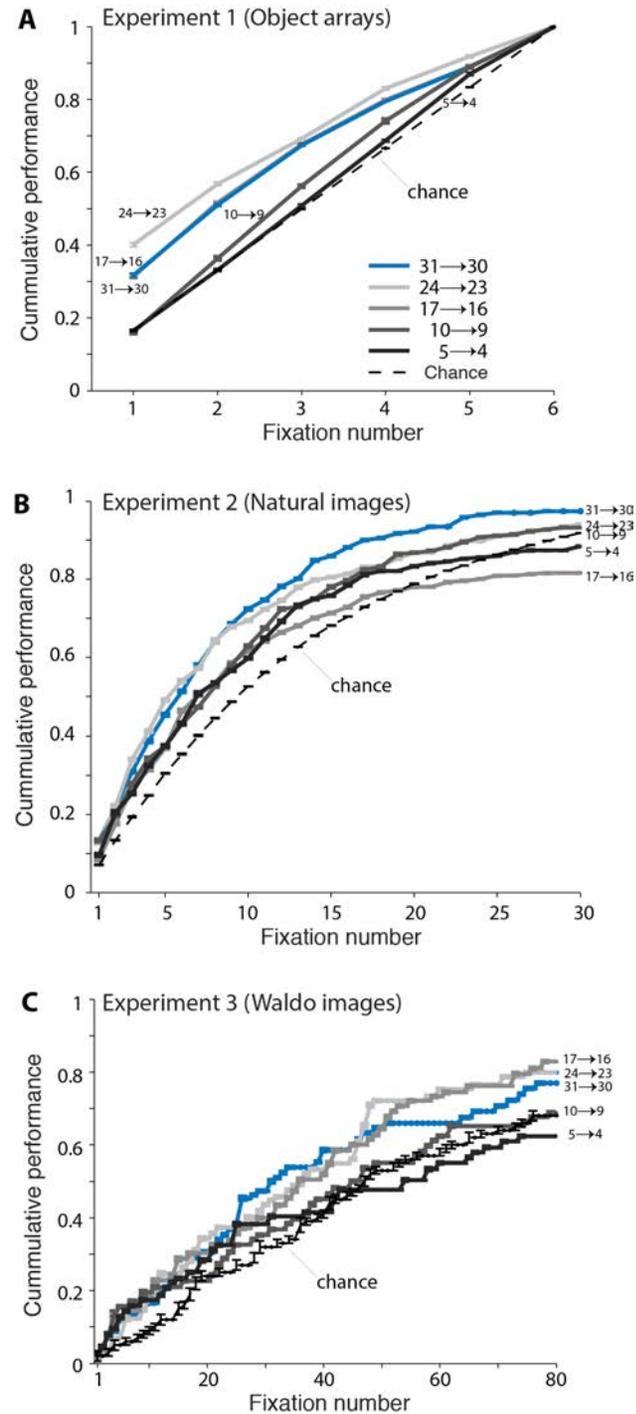

**Figure S13. Alternative IVSN models with top-down modulation at different levels of the hierarchy.** The plots follow the format of **Figures 3E**, **4E** and **5E** and the blue line is copied from those figures for comparison purposes. The curves with different shades of gray show models where top-down modulation is applied at different levels of the ventral stream hierarchy.

The performance of all models was statistically different from chance (p<0.01), except $IVSN_{5 \rightarrow 4}$ in Experiment 1 (p=0.39).

# Figure S14

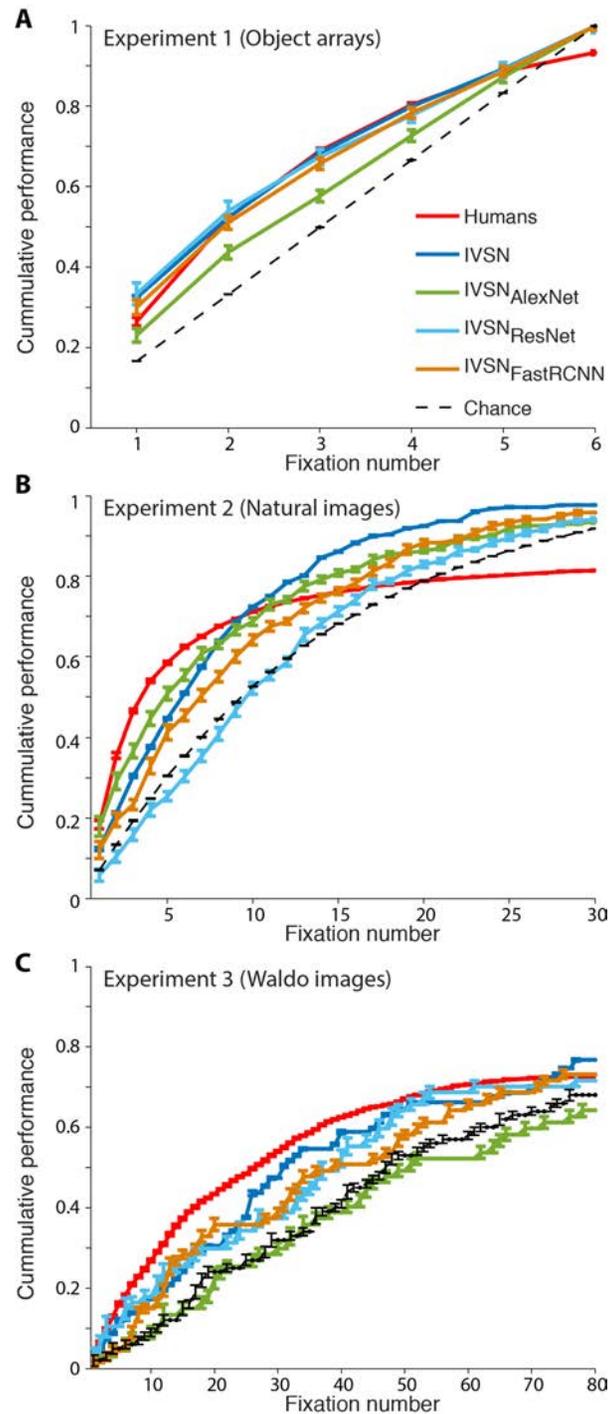

**Figure S14. Variations on the model with different ventral visual cortex modules show similar performance.** The format and conventions for this figure follow those in **Fig. 3E**. The IVSN model performance and chance levels are copied from **Figs. 3E**, **4E** and **5E** for comparison purposes. The other colors denote different models where the ventral visual cortex module in Fig. 2B was replaced by the AlexNet architecture (green), the ResNet architecture (light blue) or the FastRCNN architecture (orange). See Methods for references to these different architectures. The rest of the model remained the same.

The performance of all models was statistically different from chance ($p<0.0006$).

# Table S1

| Random | Pixel | ResNet | AlexNet | VGG16 | VGG19 |
|--------|-------|--------|---------|-------|-------|
| 0.17   | 0.21  | 0.21   | 0.22    | 0.21  | 0.21  |

**Table S1**: Category classification performance on 2000 images from 6 selected categories in Experiment 1 using various models based on "low-level" features: pixels, features from first convolution block in ResNet, Alexnet, VGG16, and VGG19 models (**Methods**). Random indicates performance obtained by selecting one of the 6 categories at random.